\documentclass{article}          % twocolumn
\usepackage{graphicx}
\usepackage{enumitem}
\usepackage{amsmath}
\usepackage{multirow}
\usepackage{amssymb}
\usepackage{subfig}
\usepackage{textcomp}
\usepackage{xcolor}

\begin{document}

\title{Swarm robotics in wireless distributed protocol design for coordinating
	robots involved in cooperative tasks
}

\author{Floriano De Rango\textsuperscript{1},
       Nunzia Palmieri\textsuperscript{1}, Xin-She Yang\textsuperscript{2},
       Salvatore Marano\textsuperscript {1} \\[10pt]
1) Dept. Computer Engineering, Modeling, Electronics and System Science,\\
	University of Calabria, Italy \\
2) Department of Design Engineering and Mathematics, \\
School of Science and Technology,  Middlesex University London, UK
}

\date{ {\bf Citation Details}:  F. De Rango, N. Palmieri, X. S. Yang, S. Marano, \\
Soft Computing, (first published online, Sept 2017).
https://doi.org/10.1007/s00500-017-2819-9 }


\maketitle

\begin{abstract}
The mine detection in an unexplored area is an optimization problem where multiple mines, randomly distributed throughout an area, need to be discovered and disarmed in a minimum amount of time. We propose a strategy to explore an unknown area, using a stigmergy approach based on ants behavior, and a novel swarm based protocol to recruit and coordinate robots for disarming the mines cooperatively. Simulation tests are presented to show the effectiveness of our proposed Ant-based Task Robot Coordination (ATRC) with only the exploration task and with both exploration and recruiting strategies. Multiple minimization objectives have been considered: the robots' recruiting time and the overall area exploration time. We discuss, through simulation, different cases under different network and field conditions, performed by the robots. The results have shown that the proposed decentralized approaches enable the swarm of robots to perform cooperative tasks intelligently without any central control.

\end{abstract}

\section{Introduction}
\label{intro}
Swarm Robotics (SR) is the study of robotic systems consisting of a large group of relatively small and simple robots that interact and cooperate with each other in order to jointly solve tasks that are beyond their own individual
capabilities. SR is becoming an emerging research area in recent years and it, mostly, inherits the inspiration from decentralized self-organizing biological systems and from the collective behavior of social insects \cite{Ref1} \cite{Ref2}.
The most important element of a multi-robot system is the ability of several individual robots to work cooperatively. By working together, the robots can
complete tasks that a single robot is incapable of accomplishing. For these reasons, multi-robot systems are applied in many engineering problems such as rescue missions, mine detection, surveillance and problems in various domains.

In this paper, we study the mine detection problem in an unknown area. It is well known that landmines are one of the biggest problems that nowadays affect many countries in the world. Such mines can remain active for years after the end of a terrible conflict and thus pose a major problem causing serious restraint and delay on post-conflict reconstruction. Despite international efforts to ban the production and use of landmines, the situation continues to deteriorate with landmines being laid about twenty times faster than they are currently being cleared.

Current technology suggests that robots could be used instead of humans to perform the demining task as minefields are dangerous to humans; thus a robotic solution allows human operators to be physically removed from the hazardous area \cite{Ref3}, \cite{Ref4}.
In this work, we focus more on the coordination of the robots to accomplish the mission than the issue of disarming physically the mines. For this purpose, we have
proposed and applied some techniques inherit from Swarm Intelligent.
Moreover, it is supposed that the robots have a number of attributes such as avoiding interference with each other,  having sensorial capabilities, sharing the workload by providing information via different sensors or wireless networks, having systems that allow the identification and disarming of mines.
More specifically, the task tackled in this paper involves three broad challenges:
\begin{enumerate}
\item Exploring unknown area: discovering of the
unknown space in the minimum time, avoiding passing more times on the same previously traversed cells;
\item Self-organization: robots can perform this task in an efficient manner through appropriate sensors on board and able to perceive the environment;
\item Recruitment task: it is the cooperative work to coordinate the robots after the detection of one or more mines and disarm them cooperatively.
\end{enumerate}

The main purpose of this paper is the presentation of an Ant based algorithm to jointly explore an unknown area and perform a recruitment/disarming task in order to analyze performance in terms of overall completion time and communication traffic to make the system highly efficient.
The objective is to find and disarm all mines and to explore all area (this last condition assures that all mines could be correctly detected in the unknown area). Our approaches are inspired by  pheromone-mediated navigation of ants and we use a direct and direct
communication mechanism for the coordination of the swarm. Through simulation, we show how this system is  able to explore unknown area in efficient manner helping the recruitment phase.

Basically, the mission is divided into two major phases: exploring and recruitment. In the exploration task, robots need to choose the direction where they will move, according to what they sense in the environment and according to the ACO based algorithm. We are interested in approaching the problem for a large group of robots following the swarm robotics principles, where the cooperation of the robots is performed, similarly in  the insect world, by an indirect communication between agents through sensing of a chemical substance (pheromone) that attract other robots in particular directions \cite{Ref5}. In our proposal, the collaborative behavior of the robots is based on the repelling anti-pheromone that means the robots try to distribute them in different regions of the area,
minimizing potentially the time.

 When one or more robots detect a mine, the recruitment process can start. In this case we propose two mechanisms. The first tries inspiration, again, from Ant Colony and use,  in this case, an attraction pheromone signaling in order to attract in the mine’s location the needed robots to perform the disarming process. The other approach uses WIFI model to communicate with the others. We propose a bio-inspired wireless distributed protocol to recruit the necessary robot in the mines location trying to reduce global communication traffic.

The paper is organized as follows: related work is presented in Section 2; problem statement and formulation are presented in Section 3. Anti-pheromone based algorithm for the exploration is described in Section 4; the attractive pheromone and the coordination protocol for the recruitment and disarming issue are presented in Section 5. Finally the performance evaluation and the conclusions are summarized in Section 6 and Section 7, respectively.

\section{Related Work}
\label{works}
\subsection{Multi-robot exploration}
Multi-robot exploration has received much attention in the research community.
The unknown area exploration should not lead to an overlapping in robots movements and ideally, the robots should complete the exploration of the area with the minimum amount of the time. The overlapped area can occur when a location has been visited by one of the robots and it is visited again by the same or different robots of the team. Many approaches have been proposed for exploring unknown environments with a team of mobile robots.

Some exploration plans in the context of mapping are usually constructed without using environmental and/or boundary information. One of the well-known techniques is frontier-based exploration, which was proposed by Yamauchi
\cite{Ref6}. In this approach, these robots act independently and make probabilistic judgements regarding frontiers areas of unexplored space in an environment. The environment is decomposed into cells with each cell being
represented by a probability value, and can be classified as either free, occupied or unknown. Using this representation a robot can reach an unexplored zone by means of navigating to the frontier cells that separate the free cells
from the unknown cells. However, other authors use different representations, and thus they identify the unexplored regions in different ways like expressed in \cite{Ref7}, \cite{Ref8}.
On the other hand, some researchers are focusing on the exploration by using knowledge about environmental boundary information, see \cite{Ref9}, \cite{Ref10}. The authors assumed that they already had the information of all obstacles. Therefore, when the robot encountered an obstacle, it could immediately
grasp the obstacle. However, this is not practical in real-world applications considering the unknown area.
Other approaches \cite{Ref11}, \cite{Ref12} coordinated the robots by means of
dividing the environment into as many disjoint regions as available robots and assigning a different region to each robot.
Tree-cover algorithms, instead, used a pre-calculated spanning-tree to direct the exploration effort and distribute it among the agents. These algorithms required a priori knowledge of the environment. A typical example is the
Multi-Robot Forest Coverage (MFC) algorithm, described in \cite{Ref13}  and Multirobot Spanning Tree Coverage (MSTC) algorithm proposed by Hazon \cite{Ref14}.

In real scenarios, we always have some uncertainty, so bio-inspired techniques have recently gained importance in computing due to the need for flexible, adaptable ways of solving engineering problems. Within the context of swarm robotics, most works on cooperative exploration are based on biologically behaviour and indirect stigmergic communication (rather than on local information, which can be applied to systems related to GPS, maps, wireless communications). This approach is typically inspired by the behaviour of certain types of animals and insects, like the ants, that use chemical
substances known as pheromone to induce behavioral changes in other members of the same species. Previous work on pheromone signalling in robotics has been used for this issue proposed in \cite{Ref15},  \cite{Ref16},  \cite{Ref17} \cite{Ref18}, \cite{Ref19}.

\subsection{Bio-inspired Self-Coordination of Multi-robot Systems}

Coordination of multi-robot has been extensively studied in the scientific literature due to its real-world applications including aggregation, pattern formation, cooperative mapping, and foraging. All of these problems consist of
multiple robots making decisions autonomously based on their local interactions with other robots and environments.
For sharing information and accomplishing the assigned tasks, there are, basically, three ways of information sharing in the swarm: direct communication (wireless, GPS), communication through environment (stigmergy) and sensing.

 More than one type of interaction can be used in one swarm; for instance, each robot senses the environment and communicates with their neighbour. In \cite{Ref20}, Tan  discussed the influences of these three types of communications on the swarm performance and the impact in a behaviour of swarm.
The self-organizing properties of animal swarms have been studied for better understanding the underlying concept of decentralized decision-making in nature, but it also gives a new approach in applications to multi-agent
system engineering and robotics. Bio-inspired approaches have been proposed for multi-robot division of labour in applications such as exploration and path formation as described in  \cite{Ref21},  \cite{Ref22}, \cite{Ref23}; cooperative transport or garbage \cite{Ref24}; inspection \cite{Ref25} and cooperation  \cite{Ref26}.
Other approaches used a direct communication among the member of the swarm. For example, Ants based routing is gaining more popularity because of its adaptive and dynamic nature and these algorithms consist in the continual acquisition of routing information through path sampling and discovery using small control packets called artificial ants. Some examples are: AntHocNet proposed by Di Caro et al. in \cite{Ref27}, Ant-Colony Based Routing Algorithm (ARA) described by Bouazizi in \cite{Ref28}. The probabilistic emergent routing algorithm (PERA) \cite{Ref29} has been proposed in which the routing table stores the probability distribution for the neighboring nodes. Singh  \cite{Ref30} presented a detail analysis of protocols based on ant-like mobile agents. Moreover, authors proposed bio-inspired routing strategies able to
minimize the number of hops, the energy wastage, see  \cite{Ref31}  or able to combine more bio-inspired techniques in the coordination actions \cite{Ref26}.

\subsection{ Recruitment as aggregation strategy}

Recruitment task is important in order to obtain a good exploitation of resources in tasks. Traditional approaches to recruitment in multi-robot
systems mainly rely on centralised coordination and require global communication. These approaches are suitable for teams of a limited number of robots and they are not suitable for swarm robotic systems which usually
consist of a large number of relatively simple robots. For swarm robotic systems, the control is completely distributed, while coordination is based on selforganisation through local interactions. Distributed coordination is suitable for multi-robot systems under a dynamic and unknown environment due to its robustness, flexibility, and reliability.

Recruitment plays a central role in social insects such as ants, bees, and termites. The recruitment task is a particular self-organization cooperative task in which robots need to aggregate in a point in order to accomplish a task as explained in  \cite{Ref33}, \cite{Ref34}.
Other approches used chemical substances to recruit the robot for certain tasks,
which was inspired by pheromone of some species of social insects, such as ants and termites \cite{Ref26}, \cite{Ref36}. Pinciroli et al. \cite{Ref37}
tried the inspiration, instead, from cockroaches. Meng et al. \cite{Ref38}
 used Particle Swarm Optimization to allocate reasonable robots to different target blocks. Other approaches use direct communication to coordinate and complete the tasks using wireless medium for communication (MANET) such as bio-inspired algorithms. An example was proposed in \cite{Ref32}.

The main contributions of this work in comparison with the literature are listed as follows:
\begin{enumerate}
\item Mathematical formulation of a multi-objective optimization
problem  accounting multiple tasks in the robot
coordination.
\item Design of swarm-based strategies where spatial and
time pheromone dispersion is applied in order to
carry out exploration and recruiting tasks (two joint
tasks).
\item Design of a protocol where the data exchanges are
balanced with stigmergy in order to assure
scalability in the robots communication and in order
to scale well in the problem complexity.
\end{enumerate}

\section{Description of the problem}
In our collective task, there are many mines randomly distributed in an unknown area. The robots should find the mine first, and then remove them. But, treating a mine is to complex by one robot, so multiple robots need to work
together. In this paper, swarm intelligence based algorithms have been proposed to search the mines and remove them. The completion time of the mission occurs when all area is explored and all mines are detected and disarmed.
Though this is a potentially NP hard problem, the objective of this  study is to develop a distributed technique for multi-robot systems in order that the robots can complete the mission as quickly as possible.

There are some assumptions for the problem that are divided into two parts: the geometry of the environment and the characteristics and capabilities of the mobile robots.
Let A be the robot’s 2-D working field, in which are distributed a finite number of static obstacles. Obstacle cells are inaccessible to the robot and impenetrable to the sensors. Let A be discretized into a grid with $m \times n$ cells.
Establish a Cartesian coordinate system which takes the upper left corner of A as the origin. Each cell has its own definite coordinate that can be represented by two coordinates (i,j), where i and j are two nonnegative
integers. At each step, the robot’s state can be represented by its location (i,j). In the area there are T stationary targets that are mines. Each target is located in a cell with coordinates (i,j). For example, T = {(0,0),(7,8),(20,6)} indicates that there are 3 mines in the area with coordinates (0,0), (7,8) and (20,6).

\begin{figure}
	\centering
	\includegraphics[scale=0.3]{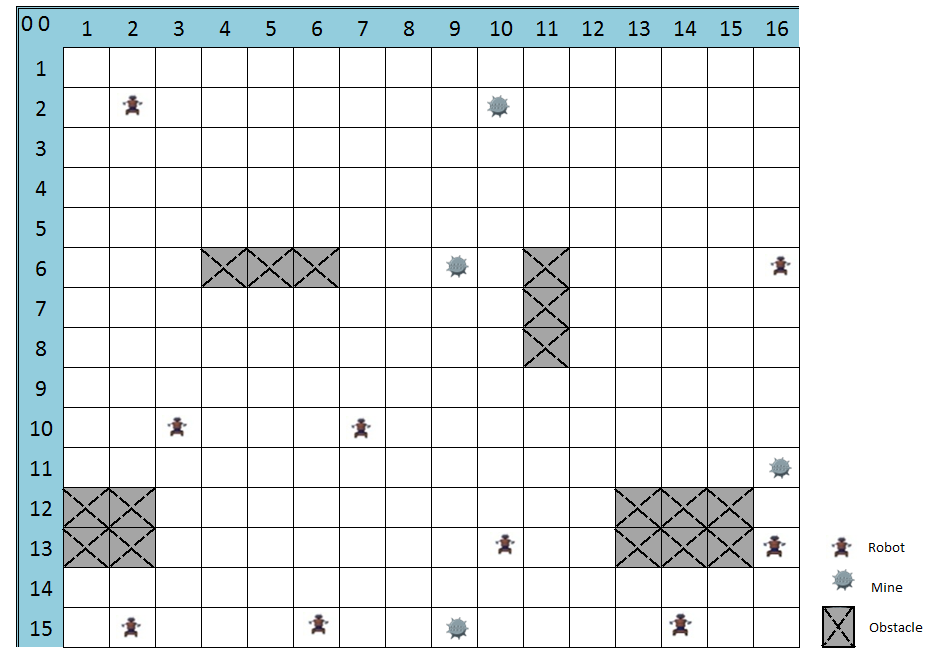}
	\caption{A representation of the considered environment.}
	\label{fig:GridArea}
\end{figure}

All robots do not have any prior
information about the location of the mines so they need to explore  the whole environment. Once a mine is detected by a robot, the recruitment process is carried out. As far as the characterization of the robots is concerned,
we assume that they live in a discrete-time domain and they can move on a cell by cell, that is, one cell at a time. They can visit all cells in the area except that the position is  occupied by an obstacles or another robot. They have limited computing and memory capacities, but not limited to motion, sensing, communication and computation. They are capable of discovering and partially executing the tasks. However, for the sake of the simplicity, the robots have a simple set of common reactive behaviour that can enable them
to avoid the obstacles and recognize the other robots in order to accomplish the mission together. The robots, at the beginning, can be placed on the same initial cell or can be randomly distributed on the grid area.
We assume that a robot uses 45\textdegree{}  as the unit for turning, since we only allow the robot to move from one cell to one of its eight neighbors, if all cells are free. The robot can have just local information about the others (neighbors
robots) in order to provide a scalable strategy. It is assumed that each robot in a cell (i,j) can move just in the neighbor cells through discrete movements Fig. \ref{fig:GridArea}.

\begin{figure}
	\centering
	\includegraphics[scale=0.3]{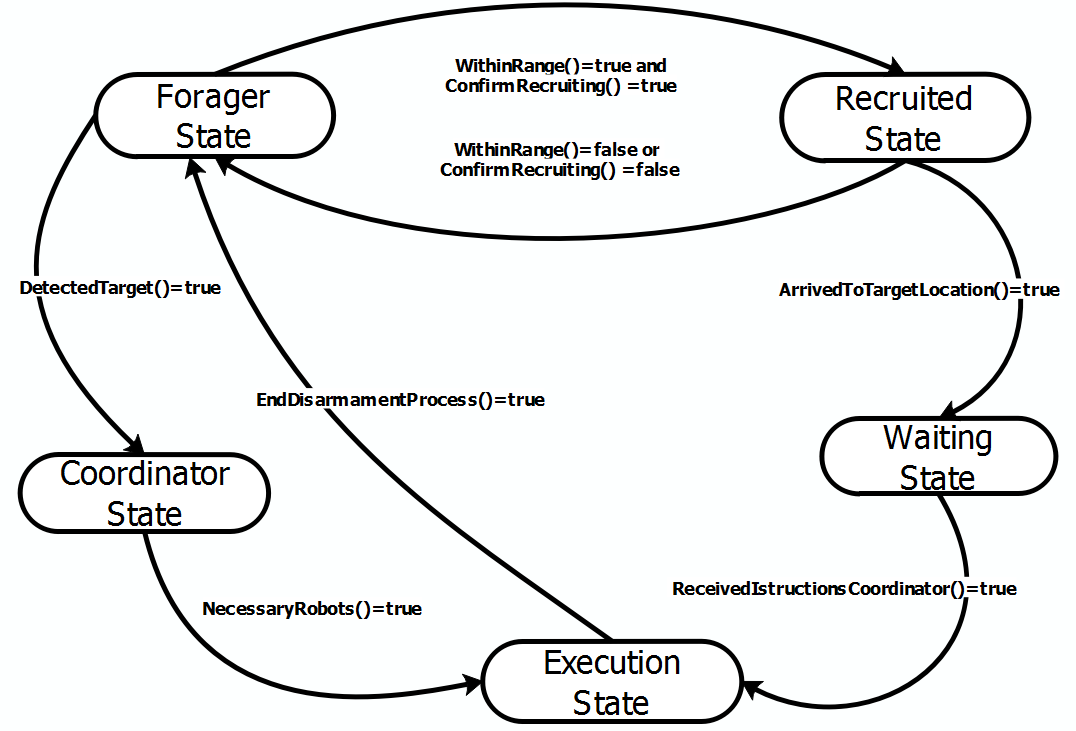}
	\caption{State transition logic for a  robot at each time step.}
	\label{fig:StateTransition}
\end{figure}

We assume that the robots are equipped with proper
sensors to perceive, leave the pheromone and detect the
mines. During the exploration task, they can leave the
pheromone in the cell and it propagates until a certain
distance. A mine is detected by a robot when the mine
position represented by the (i,j) coordinates coincides with
the robot's (i,j) location.
The behaviour of the robots, in each state, has been
described in the Fig. \ref{fig:StateTransition} on the basis of the events that can occur. We assume that the robots switch roles within a team
to carry out the tasks encountered in the environment.

More specifically, at the beginning, when no mine is detected, each robot collects information from its immediate surrounding cells perceiving chemical substance (pheromone) by onboard sensors and uses this information to identify the direction where to move. Each robot calculates its best move in terms of next position locally according to an Ant Colony-based approach as explained below. The goal is that the robots should explore the undetected
sub-areas as much as possible in order to speed up the task. This state is named the Forager State and it is the initial state for each robot.

Once a robot discovers a target by itself, it will switch to a Coordinator State. Each coordinator robot is responsible for handling the disarmament process of the discovered target and for the recruitment of the others. The recruiting process ends when the predefined
number of necessary robots ($R_{\min}$) have arrived at the target’s location to form a coalition team. Then, the accumulated robots work together as a group, performing the disarmament task.

When a robot (say, $k$) receives one or more request  by coordinator robots, it switches to the Recruited State. Then, the robot will make the decision about where to move and what target to perform. A key aspect of this state is that the robots react to events that occur. Unlike common approaches, they could change the decisions taken previously during the iterations. For example, for a certain type of mission, it is possible to meet a target or receive different requests, while reaching another target in response to a recruitment process, thus
reconsidering the choice of the target to be handled. Moreover, the decision can be to restart to explore the area since the movements are too far from the target’s location.
When a recruited robot, once it reaches the target’s location, it will wait until the other
needed robots have arrived and thus enter into the waiting mode. This state is called the
Waiting State.
Finally, once the required robots reach the target’s location, the group as a whole is
involved in the disarming process and they will perform, for a fixed amount of time, some
actions to deal with the targets properly. This state is the Execution State.

\subsection{Mathematical Model}

\label{MathematicalModel}
In order to describe the proposed system as proper mathematical models, it is useful to introduce the following notations and definitions:
\begin{itemize}
	\renewcommand{\labelitemi}{\textbullet}
	\item $A$: operational area, discretized as a grid map and $A \subset\mathbb{R}^2$
	\item 	$R$ : set of  robots
	\item $N^R$ : number of robots $N^R$ = $|$R$|$
	\item $R_{\min}$ = number of robots needed to deal with a target
	\item $T$: set of mines
	\item $N^T$: number of mines,  $N^T$ = $|$T$|$
\end{itemize}

Two main decisions have to be modelled properly. On the one hand, the position expressed by the coordinates where each robot $k$ $\in$ $R$ should be located at each step. On the other hand, given a robot $k$ and a found mine $z$ , it has to decide if it is to get involved in the manipulation process of the found target $z$.

The first decision is mathematically represented by the decision variables:	
\begin{equation}
\label{eq: ExplorationVariable}
v_{xy}^k=
\begin{cases}
1  & \text { if the robot $k$ visits the cell ($x,y$), } \\
0  & \text { otherwise.}
\end{cases}
\end{equation}

It is assumed that the time to visit a cell, denoted by $T_e$, is the same for all robots. Then the goal af an exploration task is to cover the whole area in the minimum amount of time, and thus the first objective becomes:
\begin{equation}
\textrm{minimize } \sum_{k=1}^{N^R}\sum_{x=1}^{m} \sum_{y=1}^{n}  T_{e}\ v_{xy}^k.
\end{equation}

Similarly, the following decision variables allow to model if a robot $k$ is involved in the recruitment process of the target $z$:
\begin{equation}
\label{eq: TargetVariable}
u_{z}^k=
\begin{cases}
1  & \text { if robot $k$ is involved with mine $z$, } \\
0  & \text { otherwise. }
\end{cases}
\end{equation}

When a robot has eventually detected a target, it should act as an attractor, trying to recruit the required number of robots so as to disarm the discovered mine safely and properly.

Let $T_{Start,z}^k$ be the time step at which the robot $k$ receive a help request for disarming the mine $z$ and $T_{End,z}^k$ the time step at which the robot $k$ has reached the mine $z$, then ($T_{End,z}^k$ - $T_{Start,z}^k$) is the coordination time. Thus, the objective is the minimization of the coordination time for each found mine, in order to speed up the disarming process and continue the mission effectively.
Therefore, the second objective is
\begin{equation}
\textrm{minimize } \sum_{k=1}^{N^R}\sum_{z=1}^{N^T}  (T_{End,z}^k-T_{Start,z}^k)\ u_{z}^k.
\end{equation}

\subsubsection{The Bi-Objective Optimization Problem}
The considered objective function is thus related to the minimization of the time needed to perform the overall mission. Since we have two objectives, it can be combined using the weighted sum method to convert into a single objective optimization problem. However,
because both objectives are times, and we can put the same weighting for each objective.
Thus, the optimization problem, accounting both the exploration time and the coordination time, can be mathematically stated as follows:
\begin{equation}
\label{ObjectiveFunctionTime}
\textrm{min} \
\sum_{k=1}^{N^R}\sum_{x=1}^{m} \sum_{y=1}^{n}  T_{e}  v_{xy}^k
+  \sum_{k=1}^{N^R}\sum_{z=1}^{N^T} (T_{End,z}^k - T_{Start,z}^k) u_{z}^k, 	
\end{equation}
subject to
\begin{equation}
\label{eq: ConstraintCell}
\sum_{k=1}^{N^R} v_{xy}^k \geq 1, \quad \forall\ (x,y) \in A,
\end{equation}
\begin{equation}
\label{eq: ConstraintTarget}
\sum_{k=1}^{N^R} u_{z}^k =  R_{min},  \quad \forall \ z \in T,
\end{equation}
\begin{equation}
\label{eq: DomainVariableCell}
v_{xy}^k \in \ \{ 0,1 \}, \quad  \forall \ (x,y) \in A, \  k \in R,
\end{equation}
\begin{equation}
\label{eq: DomainVariableTarget}
u_{z}^k \in \ \{ 0,1 \}, \quad \forall \ z \in T,\  k \in R.
\end{equation}
\begin{equation}
\label{eq: DomainTimeVariable}
T_e, \ T_{End,z}^k,\  T_{Start,z}^k \in \mathbb{R}, \quad \forall \ z \in T,\  k \in R.
\end{equation}

The objective function in (\ref{ObjectiveFunctionTime}) to be minimized represents the total time consumed by the swarm of robots. It depends on the time for the exploration of the area and the time for coordinating the robots involved in the disarming process of the mines. Constraint (\ref{eq: ConstraintCell}) ensures that each cell is visited at least once. Constraint (\ref{eq: ConstraintTarget}) defines that each mine $z$ must be disarmed safely by $R_{\min}$ robots.
The constraints (\ref{eq: DomainVariableCell})-(\ref{eq: DomainTimeVariable}) specify the domain of the decision variables.
The optimization problem here is intrinsically multi-objective, but it have been formulated it as a combined single objective optimization problem. Future work will focus on the extension of the current approach to the analysis of multi-objective optimization.

\section{Ant-Based Strategy for Area Exploration}
Ant colonies provide some of the richest examples for the study of collective phenomena such as collective exploration. Exploration is a very important task in nature since it allows animals to discover resources, detect the presence of potential risks, forage for food and scout for new home. Ant colonies operate without central control, coordinating their behavior through local interactions with each other. Ants perceive only local, mostly chemical and tactile cues. For a colony to monitor its environment, to detect both
resources and threats, ants must move around so that if something happens, or a food source appears, some ants are likely to be near enough to find it \cite{Ref38}.

Ant colonies, despite the simplicity of single ants, demonstrate surprisingly good results in global problem solving. Consequently, ideas borrowed from insects  and especially from ants behaviour are increasingly popular in robotics and distributed system.
Ant Colony Optimization has been developed by Dorigo \cite{Ref39}
 inspired by the natural behaviour of trail laying and following by ants.
They live in colonies and their behavior is governed by the goal of colony survival rather than being focused on the survival of individuals.
 During foraging, ants can often find shortest paths between food sources and their nest. When searching for food, ants initially explore the area surrounding their nest in a random manner. While moving, ants can leave and smell a chemical pheromone trail on the ground.
When choosing their way, they tend to choose, in probability, paths marked by strong pheromone concentrations. As soon as an ant finds a food source, it evaluates the quantity and the quality of the food and carries some of it back to the nest. During the return trip, the quantity of pheromone that an ant leaves on the ground may depend on the quantity and quality of the food. The pheromone trails will guide other ants to the food source.

The central component of an ACO algorithm is a parametrized probabilistic model, which is called the pheromone model.
Broadly speaking, the robots operate according to the following steps:
\begin{enumerate}[label=(\alph*)]
	\item The robots perceive the surrounding cells using on-board sensors.
	\item The robots compute the perceived information, in this case the concentration of pheromone, in neighbors cells.
	\item The robots decide where to go next.
	\item The robots move in their best local cell and start again from (a).
\end{enumerate}

The basic intention behind the work described here is to design a motion policy
which enables a group of robots, each equipped only with simple sensors, to efficiently explore environment eventually complex.

Broadly speaking, when the robots are exploring the area, they lay pheromone on the traversed cells and each robot uses the distribution of pheromone in its immediate vicinity to decide where to move. Like in nature, the pheromone trails change in both space and time.
The pheromone deposited by a robot on a cell  diffuses outwards cell-by-cell until a certain distance $R_s$ such that $R_s \subset A \subset\mathbb{R}^2$ and the amount of the pheromone decreases as the distance from the robot increases (see  \figurename~\ref{fig:DiffusionPheromone}).

Mathematically, the pheromone diffusion is defined as follows: consider that robot $k$ at iteration $t$ is located in a cell  of  coordinates ($x_k^t$, $y_k^t$) $\in$ $A$, then the amount of pheromone that the robot deposits at the cell $c$ of coordinates $(x,y)$ is given by:

\begin{figure}
	\includegraphics[scale=0.3]{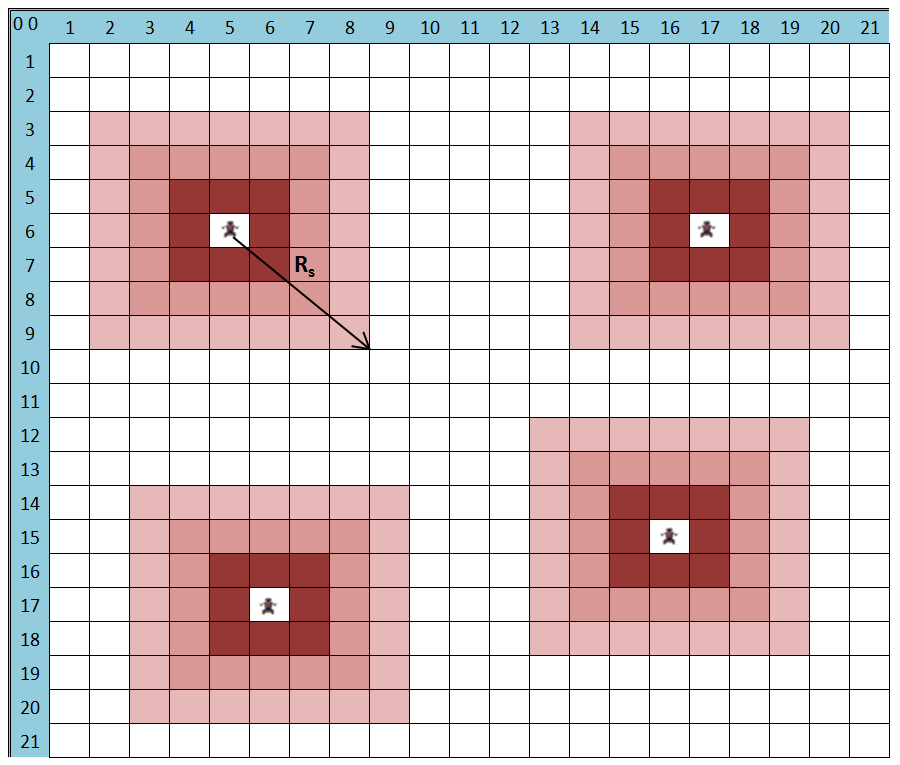}
	\caption{Robots, during the exploration, spray a chemical substance
		(pheromone) that propagates in the neighbors cells. The quantity in each
		cell depends on the distance. The cell in which the robots are moving has
		a higher quantity and decrease with the distance. The robots try to avoid
		the cells in which perceive the pheromone, in order to explore different
		regions of the area.}
	\label{fig:DiffusionPheromone}
\end{figure}

\begin{equation}
\label{eq: PheromoneCell}
\Delta\tau_{c}^{k,t} =
\begin{cases}
\Delta\tau_0\ e^\frac{- r_{kc} }{a_1} - \frac{\varepsilon}{a_2}  & \text {if  $ r_{kc}$  $\le$ $R_{s}$,   }  \\
0 & \textrm{otherwise,}
\end{cases}
\end{equation}
where $r_{kc}$ is the distance between the robot $k$ and the cell $c$ and it is defined as:
\begin{equation}
r_{kc} = \sqrt{(x_k^t-x)^2 +(y_k^t-y)^2}.
\end{equation}
This means that pheromone spreads up to a certain distance, as in the real world, after which it is no perceivable by other robots. In addition, $\Delta$$\tau_{o}$ is the quantity of pheromone sprayed in the cell where the robot $k$  is placed and it is the maximum amount of pheromone,  $\varepsilon$ is a random value (noise) so that $\varepsilon$ $\in$ $(0,1)$. Furthermore, $a_1$ and $a_2$ are two constants to reduce or increase the effect of the noise and pheromone (see Fig.~\ref{fig:DiffusionPheromoneDistance} and Fig.~\ref{fig:rhoImpact1}).
It should be noted that multiple robots can deposit pheromone in the environment at same time, then the total amount of pheromone that can be sensed in a cell $c$ depends on the contribution of many robots.

Furthermore, the deposited pheromone concentration is not fixed and evaporates with the time. The rate of evaporation of pheromone is given by $\rho$ ($0 \le \rho \le 1$ (Fig.~\ref{fig:rhoImpact}), and the total amount of pheromone evaporated in the cell $c$ at step $t$  is given by the following function:
\begin{equation}
\label{eq: pheromoneEvaporated}
\xi_{c}^t = \rho\ \tau_{c}^t,
\end{equation}
where $\tau_{c}^t$ is the total amount of the pheromone on the cell $c$ at iteration $t$.

Considering the evaporation of the pheromone and the diffusion according to the distance, the total amount of pheromone in the  cell $c$ at iteration $t$ is given by
\begin{equation}
\label{eq: totalPheromone}
\tau_{c}^t = \tau_{c}^{(t-1)} - \xi_{c}^{(t-1)} + \sum_{k=1}^{N^R} \Delta\tau_{c}^{k,t}.
\end{equation}
Each robot $k$, at each time step $t$, is placed on a particular cell $c_k^t$ that is surrounded by a set of accessible neighbor cells $N(c_k^t)$. Essentially, each robot perceives the pheromone deposited into the nearby cells, and then it chooses which cell to move to at the next step. The probability at each step $t$ for a robot $k$ of moving from cell $c_k^t$ to cell $c$  $\in$ $N(c_k^t)$ can be calculated by
\begin{equation}
\label{eq: Probabilitychhosecell}
p(c|c_k^t) = \frac{(\tau_{c} ^t)^\varphi \  (\eta_{c} ^t )^\lambda       } {\sum_{b \in N(c_k^t) }   (\tau_{b} ^t)^\varphi \  (\eta_{b} ^t )^\lambda}, \quad \forall\ c \in N(c_k^t),
\end{equation}
where $(\tau_{c} ^t)^\varphi$ is  the quantity of pheromone in the cell $c$ at iteration $t$, and  $(\eta_{c} ^t )^\lambda $ is the heuristic variable to avoid the robots being trapped in a local minimum. In addition, $\varphi$ and $\lambda$ are two constant parameters which balance the weight to be given to pheromone values and heuristic values, respectively. The robot $k$ moves into the cell that satisfies the following condition:
\begin{equation}
\label{eq: SelectedCell}
c = \arg \min [p(c|c_k^t)].
\end{equation}

In this way, the robots will prefer less frequently visited regions and more likely they will direct towards unexplored regions.

\begin{figure}
	\includegraphics[scale=0.25]{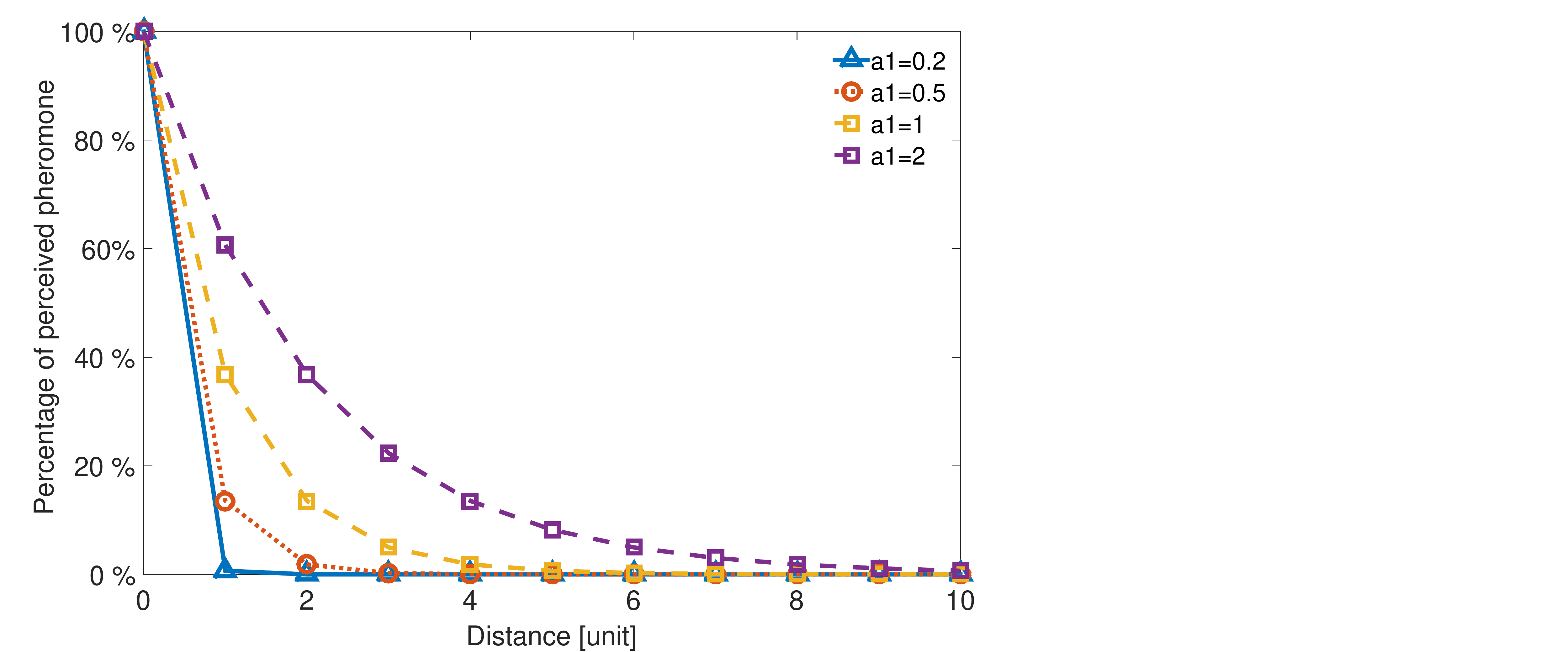}
	\caption{Trend rate of the perceived pheromone according to the distance varying $a_1$
		value.}
	\label{fig:DiffusionPheromoneDistance}
\end{figure}

 \begin{figure}
	\centering
	\includegraphics[scale=0.25]{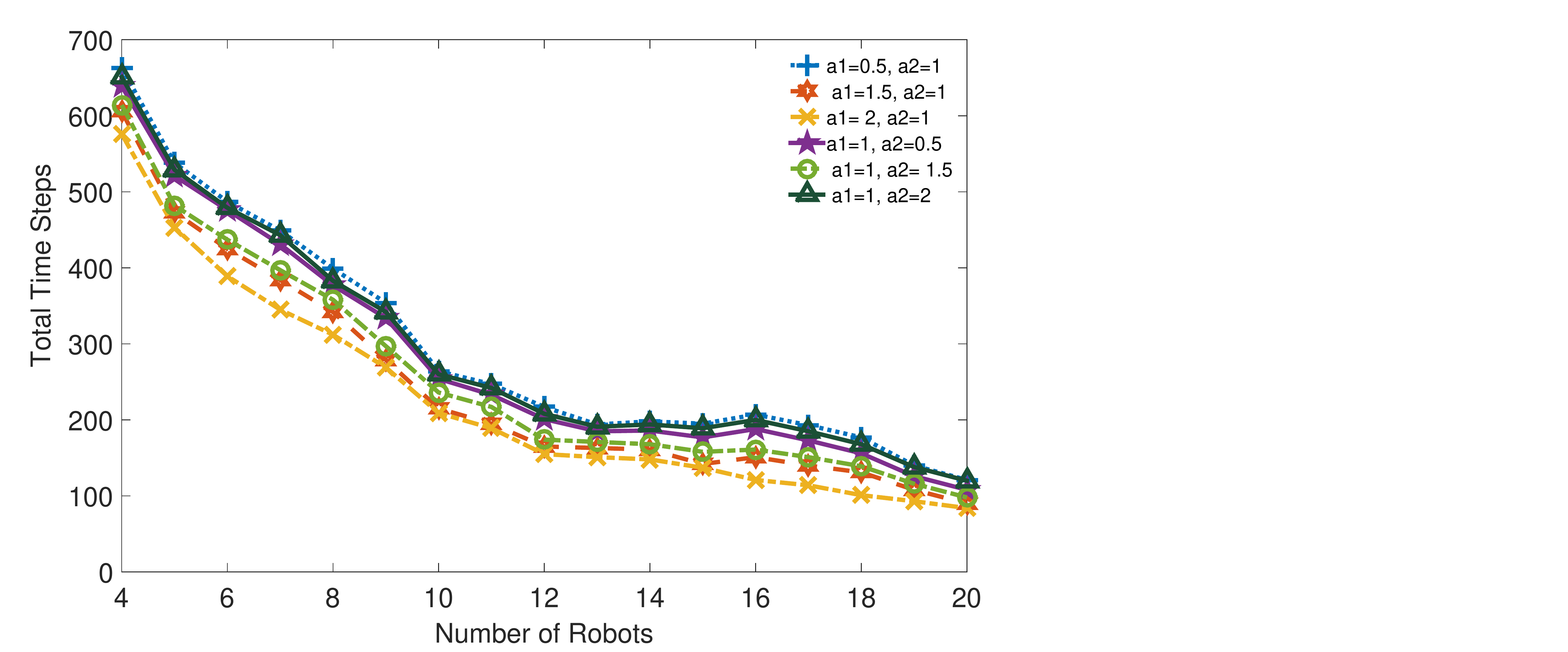}
	\caption{Impact of the $a_1$ and $a_2$ on the total time steps. }
	\label{fig:rhoImpact1}
\end{figure}

 \begin{figure}
	\centering
	\includegraphics[scale=0.25]{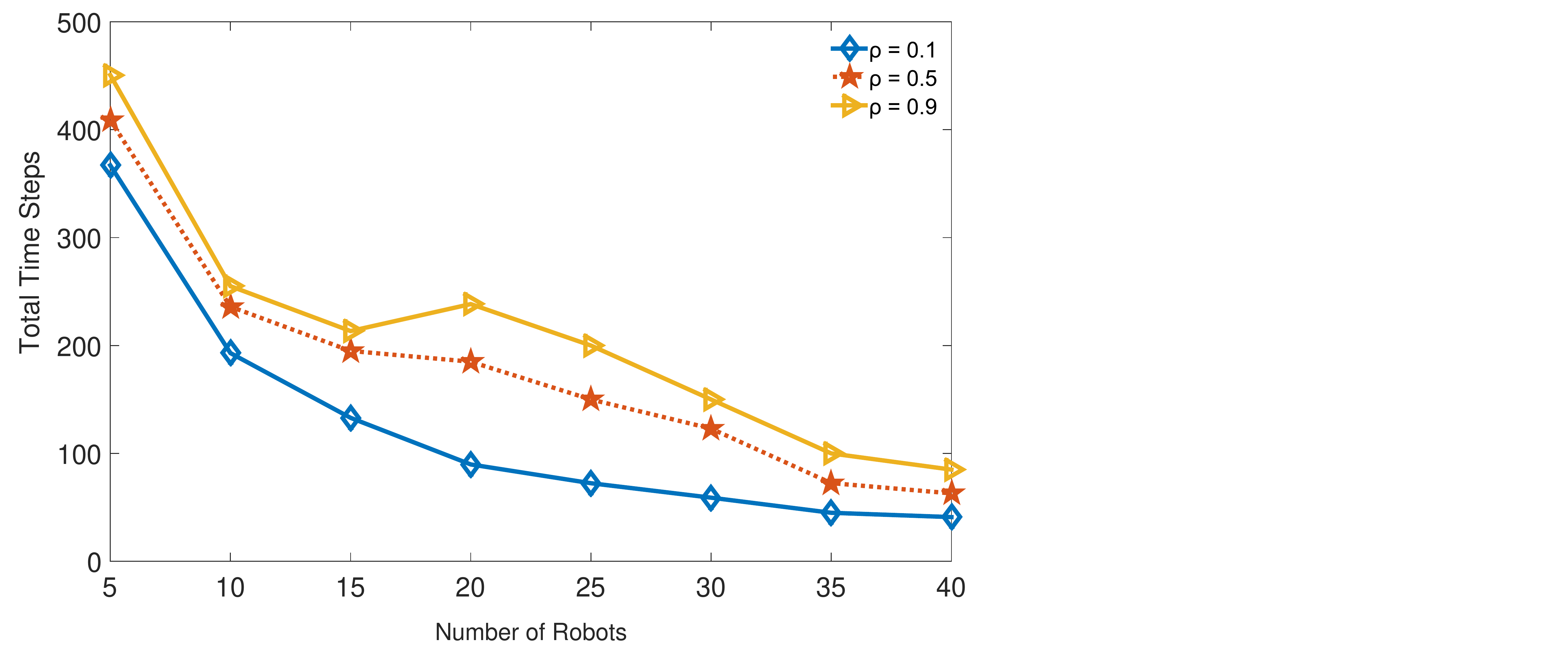}
	\caption{Impact of $\rho$ on the total time steps. }
	\label{fig:rhoImpact}
\end{figure}

\begin{figure}
	\includegraphics[scale=0.3]{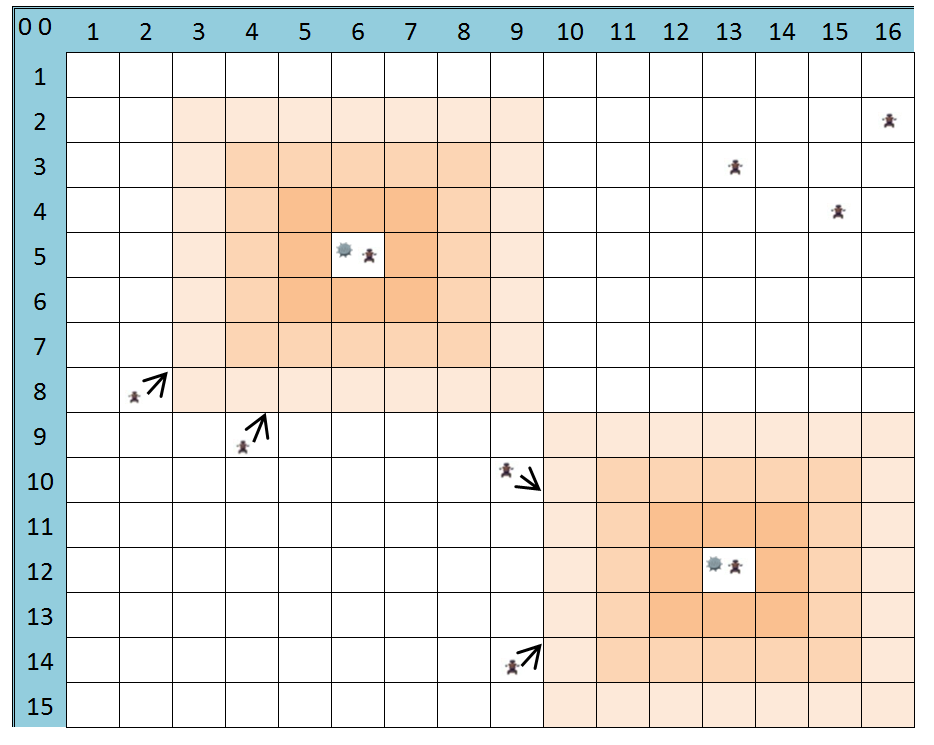}
	\caption{When the robots detect the mines, try to recruit other robots of the 		swarm spraying a pheromone. The robots that perceives the pheromone
		try to reach the mine position, preferring the cell with a higher
		concentration of pheromone that means cells probably closer to mine's
		locations. The robots, outside the pheromone range, continue the
		exploration of the area.}
	\label{fig:PheromoneMine}
\end{figure}

\section{Recruitment Strategies}

\subsection{Pheromone based Strategy for Recruitment Task}

Once a robot detects a mine by itself or receives requests
from the others, it should make the decision to search new
area or go toward a mine location to cooperate with the
others. In this case the robot that detects a mine becomes a
coordinator and would like to attract the necessary number of robots
in the mine’s location for collaborative task completion. In
our approach, the coordinator robots deposit the
pheromone, different from the previous used for exploring;
this kind of pheromone would attract other robots to guide
them into the mine’s cell. The coordinator robots continue
to spray until the necessary robots arrive into the cell
(Fig.~\ref{fig:PheromoneMine}). However, this kind of pheromone follows the same
evaporation rules explained in Section 5.
More specifically, a robot $k$ , in a cell $c_k$, that smells this kind of pheromone, chooses the next cell $c$ on the basis of the following
formula:
\begin{equation}
\label{eq: ProbabilitychhosecellMine}
p(c|c_k^t) = \frac{(\theta_{c} ^t)^\varphi \  (\eta_{c} ^t )^\lambda       } {\sum_{b \in N(c_k^t) }   (\theta_{b} ^t)^\varphi \  (\eta_{b} ^t )^\lambda}, \quad \forall\ c \in N(c_k^t),
\end{equation}
where $(\theta_{c} ^t)^\varphi$ is  the quantity of pheromone (different form the previous pheromone that has a repellent characteristic) in the cell $c$ at iteration $t$, and  $(\eta_{c} ^t )^\lambda $ is the heuristic variable to avoid the robots being trapped in a local minimum. In addition, $\varphi$ and $\lambda$ are two constant parameters which balance the weight to be given to pheromone values and heuristic values, respectively. The robot $k$ moves into the cell that satisfies the following condition:
\begin{equation}
\label{eq: SelectedCell}
c = \arg \max [p(c|c_k^t)].
\end{equation}

In this case the underlying idea was the Maximum
Pheromone Following to allow to the robots to reach the
mine’s location with a lower time.
The mechanism that uses the pheromone in the
exploration phase and in the recruiting phase is called
ATRC-ERS (Exploration and Recruiting with only
Stigmergy). It exploits just stigmergy, and the robots
change behavior from Minimum Pheromone Follower to
Maximum Pheromone Follower based on the roles that they
assume during the mission.

\subsection{Distributed Wireless Communication for Robots Coordination}

In this section an on-demand mobile ad hoc network related to the problem to form coalitions in certain locations of the area is presented. The network architecture is created once a robot detects a target in the area and from this point that initiates communication with neighbor to neighbor. The idea is to use ad hoc routing protocol to report a detected target and the robots
that wants to serve it over a MANET.
Mobile ad-hoc networks (MANETs) consist of special kind of wireless mobile nodes which form a temporary network without using any infrastructure or centralized administration.
In networks, all nodes are mobile and communicate with each other via wireless
connections. Nodes can join or leave the network at any time. There is no fixed infrastructure.
All nodes are equal and there is no centralized control or overview. There are no designated routers: all nodes can serve as routers for each other, and data packets are forwarded from node to node in a multi-hop fashion. Since in mobile ad-hoc networks there is no infrastructure support and nodes being out of
range of a source node transmitting packets; a routing procedure is always needed to find a path so as to forward the packets appropriately between the source and the destination.

Moreover, due to limited resources such as power, bandwidth, processing capability, and storage space at the nodes as well as mobility, it is important to reduce routing overheads in MANETs, while ensuring a high rate of packet delivery. Due to the dynamic nature of MANETs, route maintenance is quite a difficult task. Basically, routing is the process of choosing paths in a network along which the source can send data packets towards destination. Routing is an important aspect of network communication because the characteristics like throughput, reliability and congestion depends upon the
routing information. An ideal routing algorithm is one which is able to deliver the packet to its destination with minimum amount of delay and network overhead. The nodes update the routing tables by exchanging routing information between the other nodes in the network.

In the literature there exists a large family of ad hoc routing protocols. However, it has been found that bio-inspired approach such as ant colony optimization (ACO) algorithms can give better results as they are having characterization of Swarm Intelligence (SI) which is highly suitable for finding the adaptive routing for such type of volatile network.
ACO routing algorithms use simple agents called artificial ants which establish optimum paths between source and destination that communicate indirectly with each other by means of stigmergy. The basic idea behind ACO algorithms for routing is the acquisition of routing information through sampling of paths using small control packets, which are called ants. The ants are
generated concurrently and independently by the nodes, with the task to test a path to an assigned destination. An ant going from source node $s$ to destination node $d$ collects information about the quality of the path (e.g. end-to-end delay, number of hops, etc.), and uses this on its way back from $d$ to $s$ to update the routing information at the intermediate nodes.

The routing tables contain for each destination a vector of real-valued entries, one for
each known neighbor node. These entries are a measure of the goodness of going over
that neighbor on the way to the destination. They are termed pheromone variables, and are
continually updated according to path quality values calculated by the ants. The repeated
and concurrent generation of path-sampling ants results in the availability at each node of a
bundle of paths, each with an estimated measure of quality. In turn, the ants use the routing
tables to define which path to their destination they sample: at each node they stochastically
choose a next hop, giving higher probability to links with higher pheromone values. For this
reason, generally, the routing tables are also called pheromone tables. The routing table at
each node is organized on a perdestination basis and is of the form (Destination, Next hop,
Probability). It contains the goodness values for a particular neighbor to be selected as the
next hop for a particular destination. Further, each node also maintains a table of statistics
for each destination d to which a forward ant has been previously sent.

More specifically, the network of robots is created when one or more robots find a target.
That is, the robot that has detected a target sends announcement messages that are forwarded by the other robots so that the information about the target can spread among the swarm.

The messages that a robot can send or receive are:

\begin{enumerate}
	\item \textbf{HELLO}: Hello packets are used to notify the robot presence in its transmission range to other robots. A HELLO packet contains the ID of the sending robot. When a robot receives this packet becomes aware of the presence of another robot in its range and it writes the ID in a data structure (neighbors table) which takes into account all the robots in the direct communication range. If, after a time period, it does not receive HELLO packets from other robots listed in its neighbor table, it deletes the correspondent entry line. In this way, a robot will know the robots that can be reached directly (one-hop).
	\item	\textit{Requiring Task Forward Ant} (\textbf{RT-FANT}): it is a packet sent by the robot that has detected a mine (that is the coordinator robot) to know how many robots are available to treat the mine.
	\item	\textit{Requiring Task Backward Ant}  (\textbf{RT-BANT}): it is a packet that a robot in Forager State sends as response to a RT-FANT.
	\item	\textit{Recruitment Fant}  (\textbf{R-FANT}): it is a packet sent by a coordinator, to the link from which came the higher number of RT-BANT responses; this link has a higher recruitment probability.
	
	\item	\textit{Recruitment Bant}  (\textbf{R-BANT}): it is a packet sent by a robot in response to a positive recruitment by a coordinator.
	\item	Leaving position  (\textbf{LP}): if a R-BANT, generated by a robot in response to the R-FANT, does not arrive to coordinator within a certain time (it is a timer), and in target's location has arrived the needed robots, the coordinator sends this message informing these robots to continue to explore the area or serve eventually other requests.
\end{enumerate}

In the following the actions, in terms of received packets are described, in order to deeply understand the functioning of the protocol and the difference of packets that are sent during the mission.

For the most time, a robot is in \textit{Forager} executing the exploration task. Its operations are essentially the following:

\begin{enumerate}[label=\Roman{*}., ref=(\Roman{*})]
	\item	Process packets content: when a robot receives a packet it forwards the packet to another destination.
	\item	Exploration phase according to exploration algorithm.
\end{enumerate}

\begin{figure}
	\centering
	\includegraphics[scale=0.39]{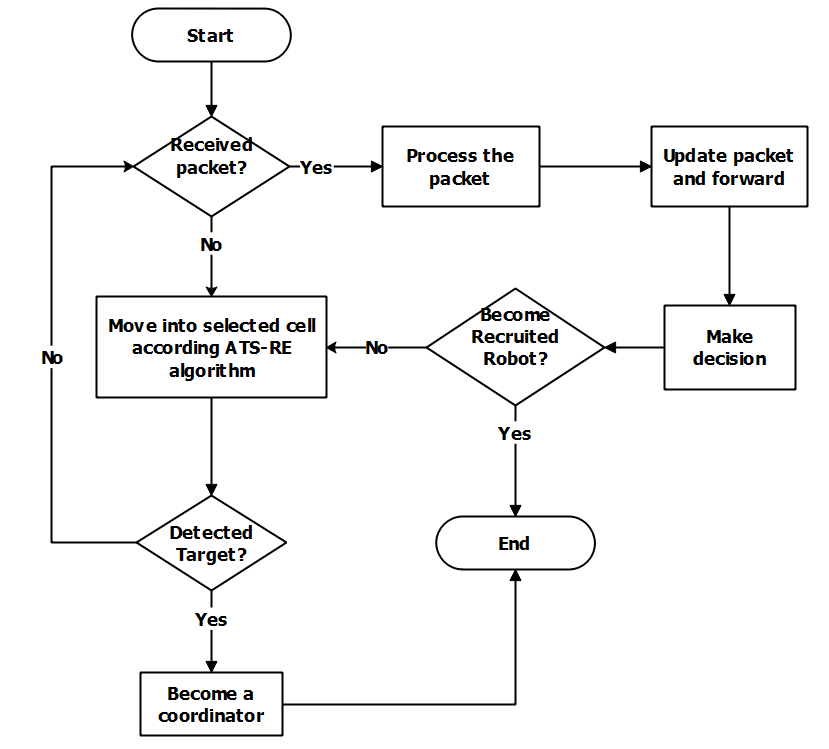}
	\caption{The Flow Chart of a Forager Robot.}
	\label{fig:FlowChartForager}
	\end{figure}

%%	\newpage
	A coordinator robot performs these operations:
	\begin{itemize}
		
		\item	FANT Generating and Forwarding: it creates and sends broadcast requests in the network; in this step the coordinator sends a  \textbf{RT-FANT} to know how many robots are, eventually, available for disarming the found target. The RT-FANT, identified by the triple (ID-Coordinator, Task-ID, ID-FANT), is sent in broadcast to all robots in the transmission range.
		\item	Set waiting timer: after sending the RT-FANT, the coordinator sets a timer to wait the RT-BANT packets sent by robots available to be recruited; after timing out it checks the number of received RT-BANT. If the coordinator does not receive enough replies, analyses the number of received replies: if it does not receive any replies it becomes a Forager, else it creates and sends a new Request Task FANT and forwards in broadcast on the network. If the coordinator has enough replies (RT-BANT) to perform the task, it creates and sends R-FANT on the link with a higher recruitment probability.
		\item	Wait incoming robots: the coordinator waits for the incoming recruited robots.
		\item	Submit disarming order: When all needed robots are recruited into the interested cell, the coordinator sends a message to announce the starting of the manipulation task of the target.
	\end{itemize}
	
	\begin{figure}
		\includegraphics[scale=0.35]{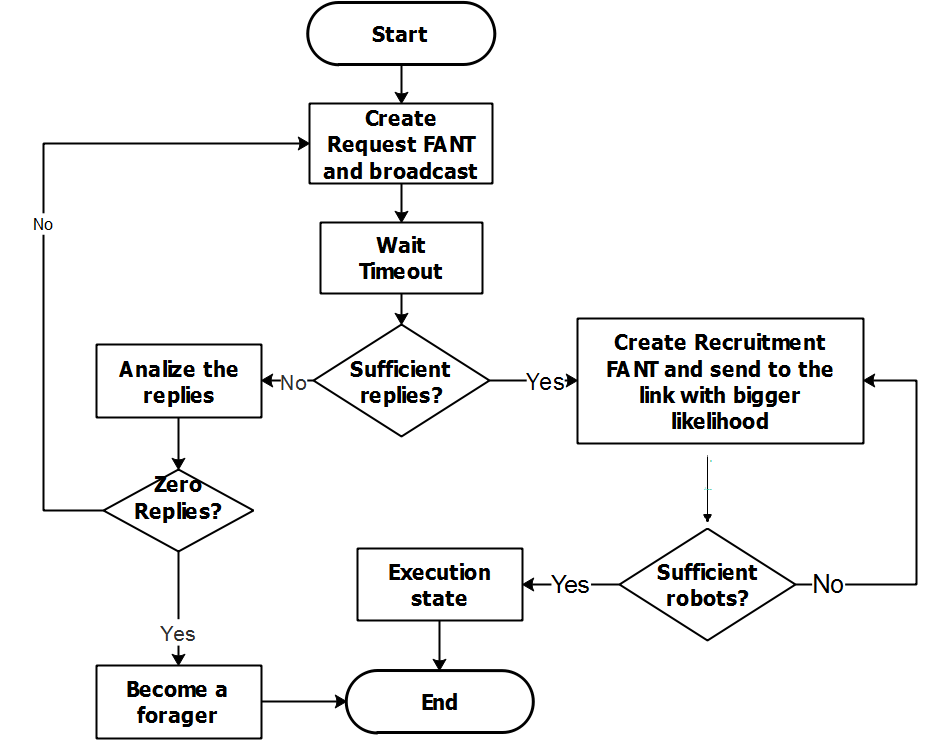}
		\caption{The flow Chart of a Coordinator Robot.}
		\label{fig:FlowChartCoordinator}
		\end {figure}

		When a robot receives a RT-FANT packet and sends a RT-BANT to the coordinator, it becomes a \textit{Recruited Robot}. Then, its task is to reach the destination cell. Essentially, the recruited robot moves into the area in order to reach the target's location.

		\begin{figure}
			\includegraphics[scale=0.33]{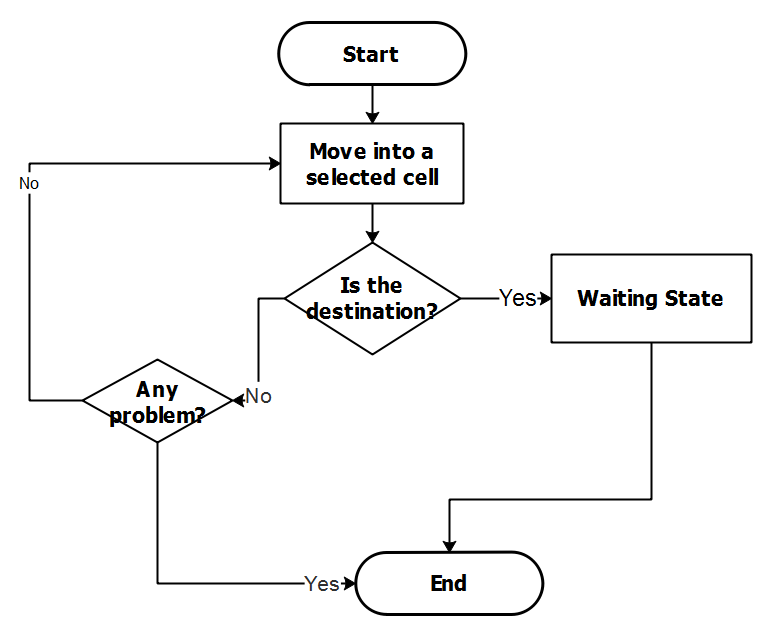}
			\caption{The Flow Chart of a Recruited Robot.}
			\label{fig:FlowChartRecruited}
			\end{figure}

			\subsection{Forwarding mechanism of FANT and BANT}
			
In the considered  problem, an Ant-based Team Robot Coordination (ATRC) protocol has been applied and it uses typically probabilistic routing tables to establish to which robots distribute the coordination tasks. This routing table is populated and updated on the basis of the packets sent from coordinators to recruited (Forward ANT: R-FANT and RT-FANT) and vice versa (Backward ANT: R-BANT and RT-BANT). To ensure that for every FANT sent on the path from the coordinator to the recruited sent back a BANT on the reverse-path forwarding to the coordinator, each node crossed by the FANT enters its ID in the packet. Once it reaches its destination a Backward ANT (BANT) response is created; in this packet the ID of crossed robots and additional information for updating the routing tables are copied. BANT follows the route tracked by FANT so it reaches the destination host (coordinator). For this behavior, the two considered packets are called Forward (FANT) and Backward ANT (BANT).

			During this discovery procedure, BANT updates the entry in the routing table of the node. The law for updating the pheromone is usually based on the path length, that is the number of hops (in terms of robots) crossed by FANT to reach the destination. The routing table in this work are not deterministic, but probabilistic.

Essentially a packet has the following fields:		
			\begin{itemize}
				\item[-] ID Coordinator: ID of the coordinator robot and it is added in a RT-FANT;
				\item [-] Task ID: it is the ID of the task requested by the coordinator. Each time the same coordinator runs different tasks this value is incremented.
				\item [-] Task Type: in this case there are three tasks (recruiting, disarming and discovery), but this field can be useful for future purpose and extensions to multiple and more complicated tasks.
				\item [-] Path Degree $p_D$: it is a weight given to a path in order to understand which route can be the best according with some specific metrics; it can affect the link selection probability for each link between the current robot and its neighbors.
			\end{itemize}
			
			The ID Coordinator, Task ID and Task Type allow the unique identification of an entry. Initially, when a RT-FANT is sent on the network, each robot receives RT-FANT and creates an entry in the routing table and sets a balanced selection probability of the neighbors. These probabilities are then updated through the response RT-BANT. Each robot that receives an RT-BANT from a particular link, updates the probability associated to that link and decreases the other link probabilities through the use of two concepts:
			
			\begin{enumerate}
				\item  Evaporation
				\item  Reinforcement.
			\end{enumerate}
			
			The evaporation is applied to all links, while reinforcement learning is applied to the link receiving the RT-BANT. The quality of a link depends on the distance of the robot that creates the RT-BANT to the destination (cell where the mine needs to be deactivated).
			In this way the probability of the link that receives the highest number of RT-BANT increases.
			Having to submit the RT-FANT in a deterministic way, a robot is able to choose the link with the highest recruitment probability.
			Also, the received R-BANT contains a recruitment task during the travelling for each link, the robot only executes the process of evaporation. This is made to improve the link selection probability, indicating a high number of robots willing to perform the task requested.

	\begin{figure}
				\centering
				\includegraphics[scale=0.32]{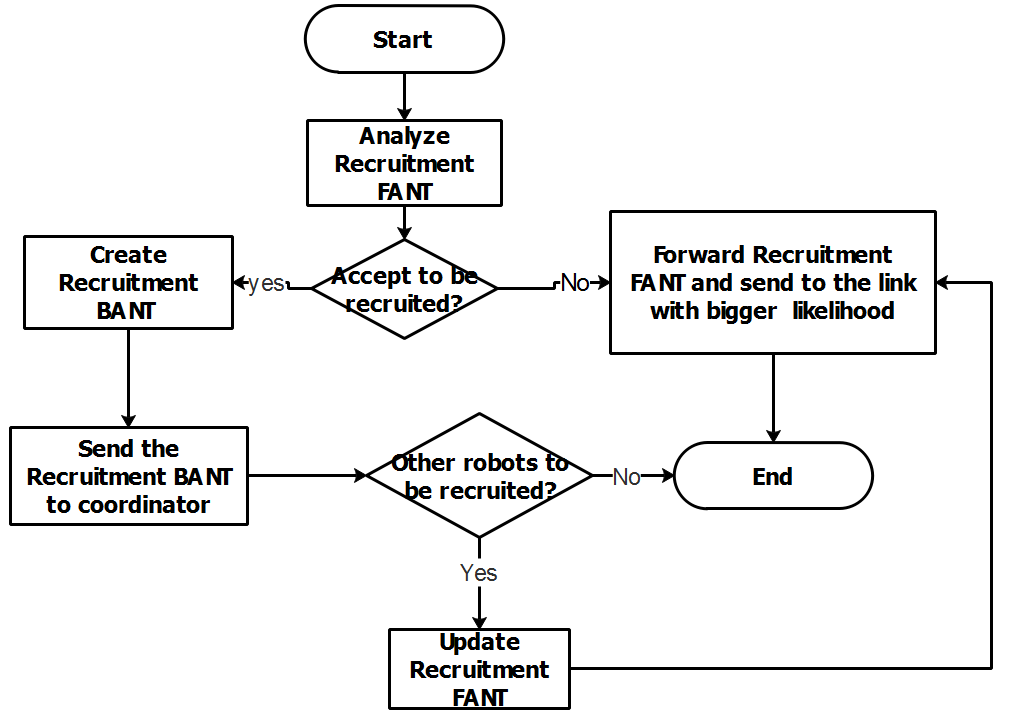}
				\caption{The Flow Chart of a Recruitment FANT and BANT.}
				\label{fig:FlowChartRecruitmentFANTBANT}
					\end{figure}

			\begin{figure}
			\centering
			\includegraphics[scale=0.43]{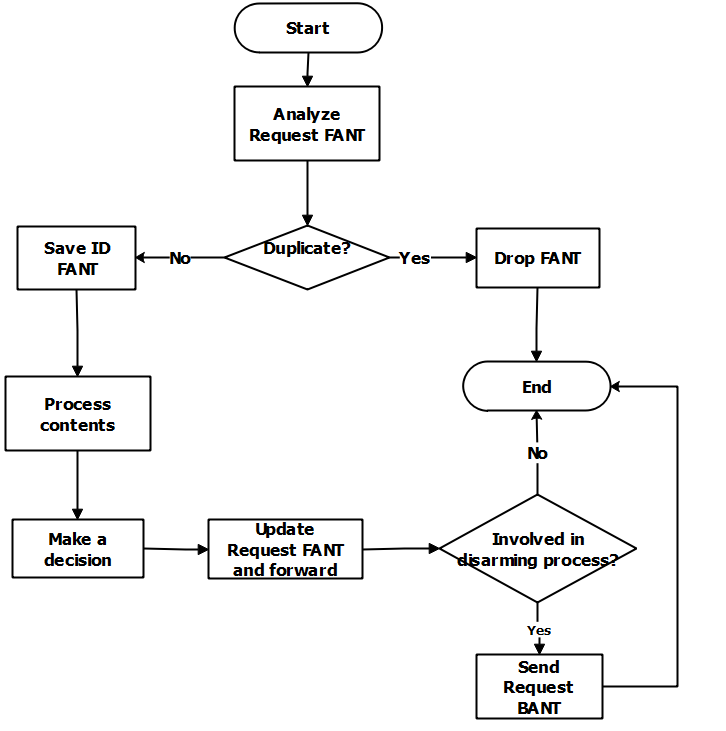}
			\caption{The Flow Chart of a Request FANT and BANT.}
			\label{fig:FlowChartRequestFANTBANT}
		\end{figure}			
					
\subsection{Task Requesting BANT and FANT Management}
					
					When the coordinator sends RT-FANT, only foragers process this packet. If the packet is received by robots that are in other state they forward in broadcast the RT-FANT.
					The forager receiving RT-FANT performs the same operations below:
	\begin{itemize}
						\item	Checking uniqueness of received FANTs: a forager, after receiving a packet containing RT-FANT, controls if it processed this packet previously. In this case the robot drops the packets and carries on its operations, otherwise it saves the ID FANT in a data structure and processes the packet content.
						\item	Process requirements: If the received RT-FANT is not duplicated, the forager checks the required characteristics. If it is able to perform the task, it controls the percentage of BANTs already forwarded to the coordinator, according with previously forwarded FANTs, and decides, in a probabilistic manner, whether to forward or not its answer. Next it creates and sends an RT-BANT to the coordinator. The forager, finally, sends the received RT-FANT in broadcast also if it is not able to perform the task.
	\end{itemize}

\subsection{ Recruitment FANT and BANT Management}
					A coordinator, after receiving enough responses by foragers, sends R-FANT on the link that has the highest success probability.
					The foragers receiving this FANT execute these operations:
					\begin{itemize}
						\item	Processing R-FANT: Initially, the forager checks if the FANT has been previously processed; in this case it discards the packet. In other case it adds its identifier in the list of crossed robots by R-FANT and then processes the recruitment request.
						\item	BANT Management: if the robot decides to participate in the disarmament of the target, it creates and sends a R-BANT to coordinator as a recruitment confirmation. The R-BANT updates the routing table of the crossed nodes.
						\item	FANT Forwarding: independently by the response of R-BANT, a forager receiving a R-FANT creates and sends new R-FANT to other robots if there is the need to recruit other robots on the link with higher recruitment probability otherwise, if itself is the last robot, it does not forward any R-FANT.
					\end{itemize}

\section{Simulation Experiments}
\begin{figure}[hp!t]
	\centering
		\includegraphics[scale=0.3]{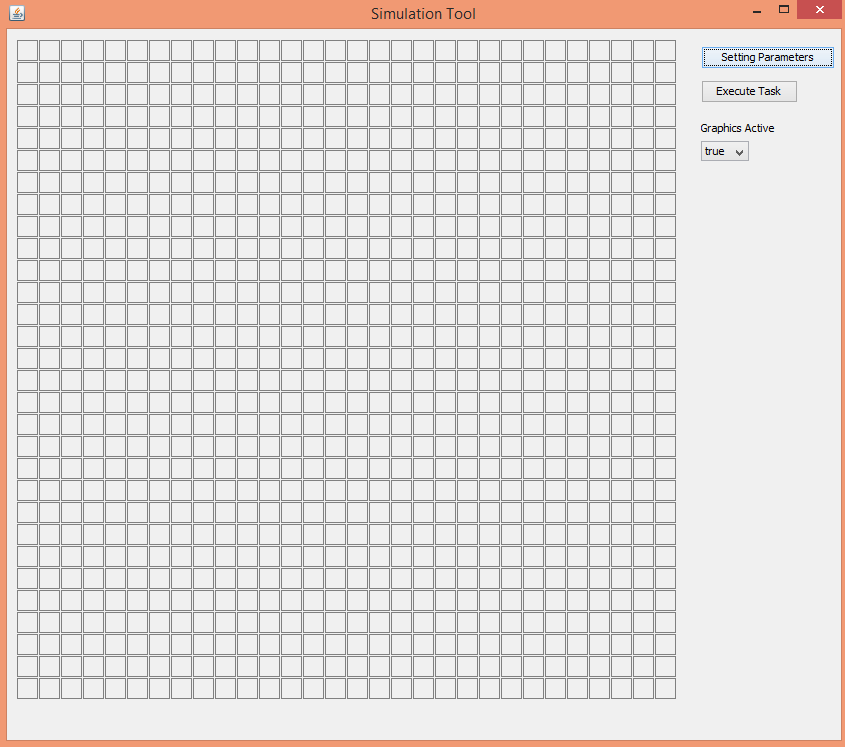}\quad
		\centering
		\includegraphics[scale=0.3]{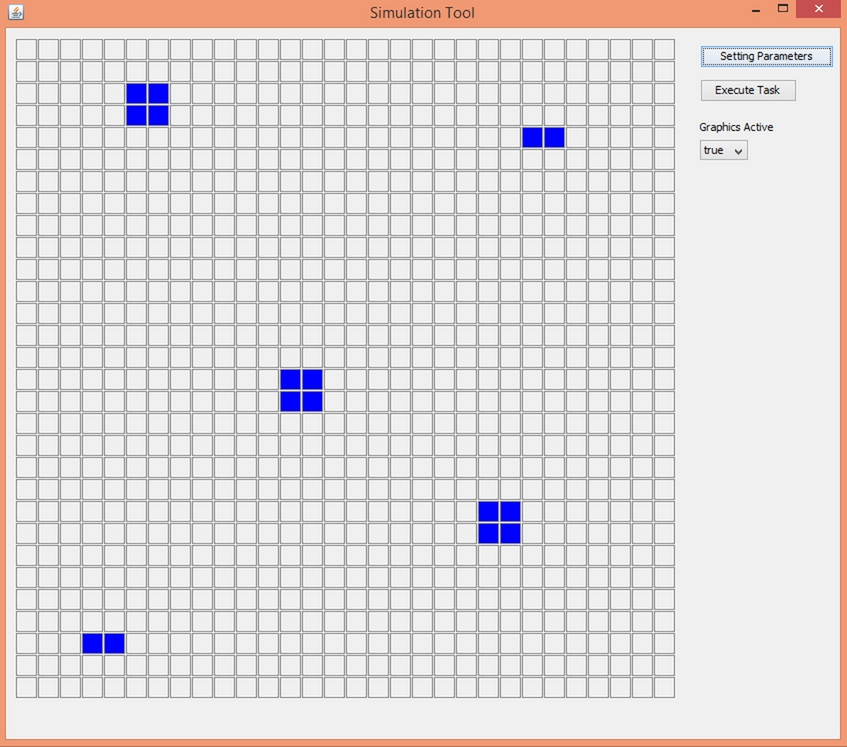}
	
	 \quad
	\centering
	\includegraphics[scale=0.35]{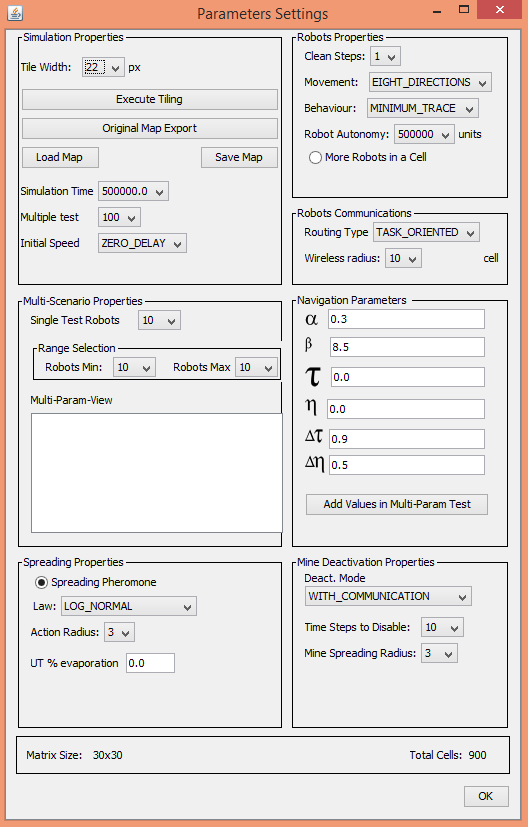}
	\caption{ Simulator Front-end (a) Environment without obstacles (b) Environment with obstacles (c) Parameters Setting. }
	\label{fig:SimulatorScreenshot}
	
\end{figure}

A set of experiments have been performed in order to show and analyze the effectiveness of the proposed approach.
For such purpose, a hand-designed simulator have been implemented in Java. This simulator was built from the start as a multi-robot simulator. It is capable of modeling motion, targets, obstacles and local communication in a discrete world, and it can be easily extended to simulate other scenarios and domains since it is generalized.
Screenshots of the simulator’s graphical output option could be seen in Fig. \ref{fig:SimulatorScreenshot}, in which the parameters, regarding both exploration and recruiting tasks are represented.

The simulations were executed varying different
parameters of the problem. We started to evaluate the Antbased
Team Robot Coordination (ATRC) with only
exploration in comparison with IAS-SS
proposed by Calvo \cite{Ref15} et al.  in an area with obstacles
and not. Later, we evaluated the performance of our
algorithm with both exploration and recruiting strategy
applying the wireless communication (ATRC-ERP) or
using just stigmergy (ATRC-ERS).

The performance metrics considered for the simulation are:
\begin{itemize}
\item Average Task Execution Time: it is the total task
execution time evaluated in terms of number of
iterations. If more tasks are considered such as
exploring, recruiting and disarming, this metric
accounts for the total average time necessary to
complete all tasks.
\item Control Overhead: it accounts for the number of
control packets such as R-FANT, R-BANT, RTFANT,
RT-BANT sent on the network to perform
the protocol operations.

\end{itemize}
In Table I the simulation parameters are shown.
We have used a minimum of 4 robots to disarm the mine (Rmin=4)
changing the number of robots in a mined region; the transmission range $R:t$=9; this value
has been fixed just to reduce the number of simulations due
to space limitations. However, the proposed approach is
general and the RW value can be also changed without
affecting the algorithm convergence and simulation trend. In addition,
we considered a grid
area without obstacles and with obstacles, varying, during
the simulation tests, the number of grid cells.

\subsection{ Stigmergy aware Space Discovery vs Protocol
aware Bio-inspired strategy}

\begin{figure}
	\centering
	\subfloat[][\emph]
	{\includegraphics[width=.50\textwidth]{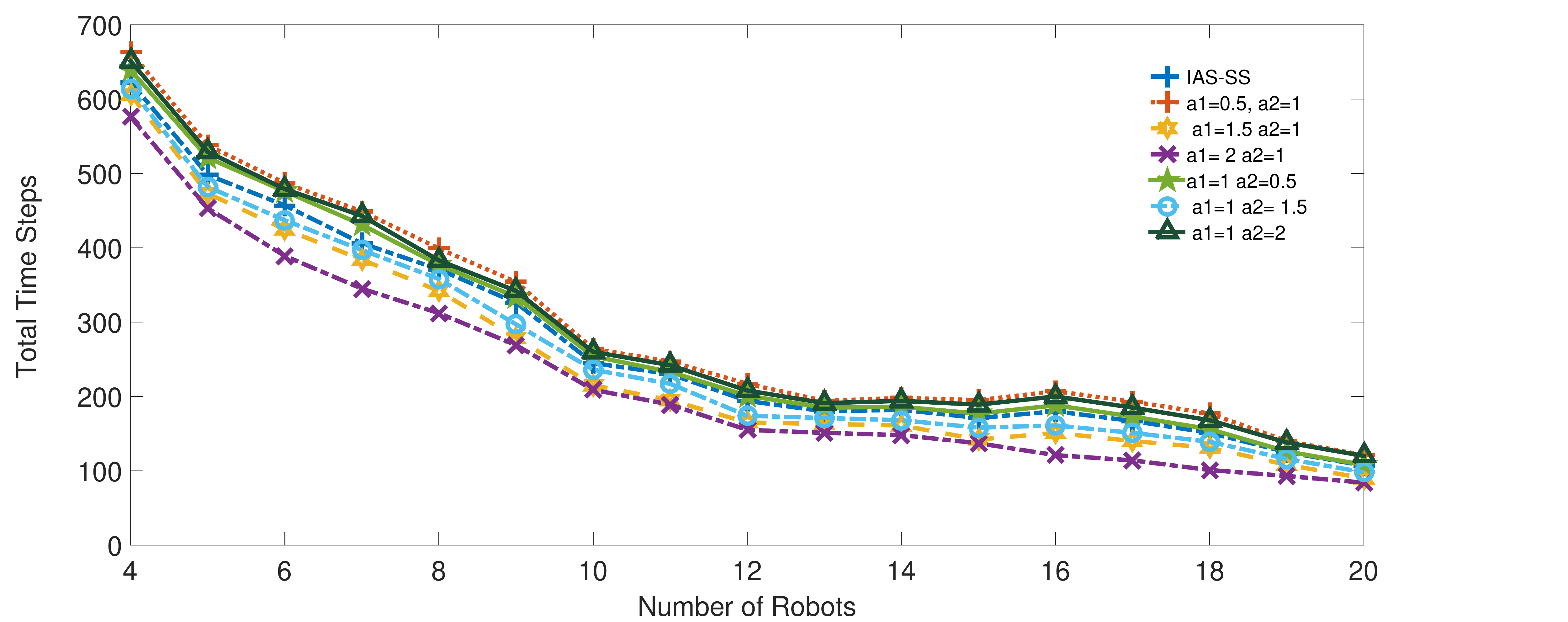}} \quad
	\subfloat[][\emph]
	{\includegraphics[width=.55\textwidth]{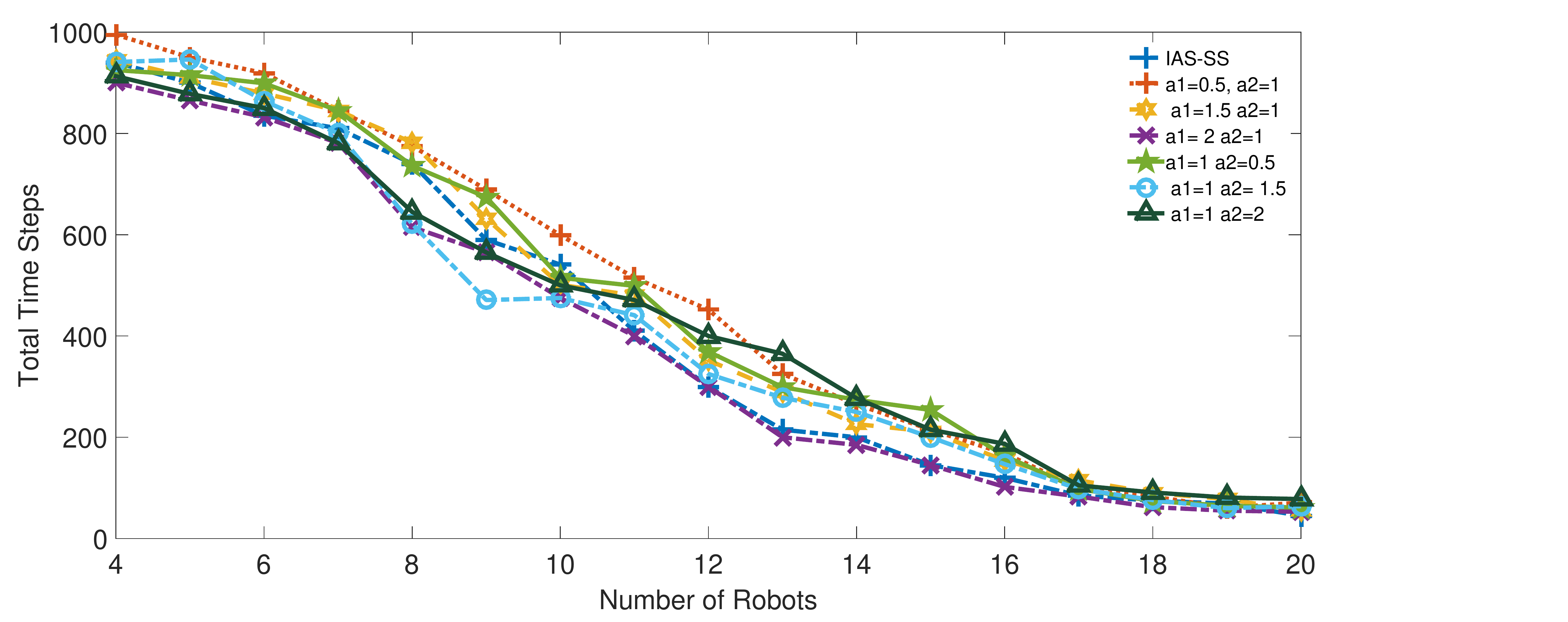}} \quad
	
		\caption{ ATR-RE vs IAS-SS  (a) Free Environment (b) Obstacle Environment.}
	\label{fig:IAS-SSvsATR-RE}
\end{figure}

We evaluate, firstly, the performance of the proposed
exploration algorithm (ATRC-OE) in comparison with
IAS-SS \cite{Ref15}. This last strategy tries
inspiration by the inverse ant-colony optimization and it
can be considered as a special case of our proposal
changing in opportuning manner the $a_1$ and $a_2$  value.
In Fig. \ref{fig:IAS-SSvsATR-RE} the performance of both strategies are
depicted varying the value of the parameters in the problem
such as $a_1$ and $a_2$.
The figure considers the total time to explore,
increasing the number of robots is shown. As we expected, a
higher number of robots reduces the cells discovery time
for both IAS-SS and ATRC-OE.

Our approach is able to
obtain a lower discovery time through the swarm based
solution. The trend is similar both in free environment and in environment with obstacles.
Generally, a higher number of robots can assure a lower
convergence time.
However, we do not need to increase a lot the number of
robots but we can stop to a minimum number after which
no more gain is obtained

\subsection{ ATRC-ERP vs ATRC-ERS Performance}
In this subsection, ATRC with exploration and
recruiting tasks has been evaluated. Two versions of the
ATRC with only the stigmergy to perform both tasks
(ATRC-ERS) and with the addition and support of the bioinspired
protocol (ATRC-ERP), such as explained in
Section 5, has been tested under different parameters
conditions in order to verify its robustness, convergence
and scalability for increasing complexity.

\begin{figure}
	\centering
	\includegraphics[scale=0.3]{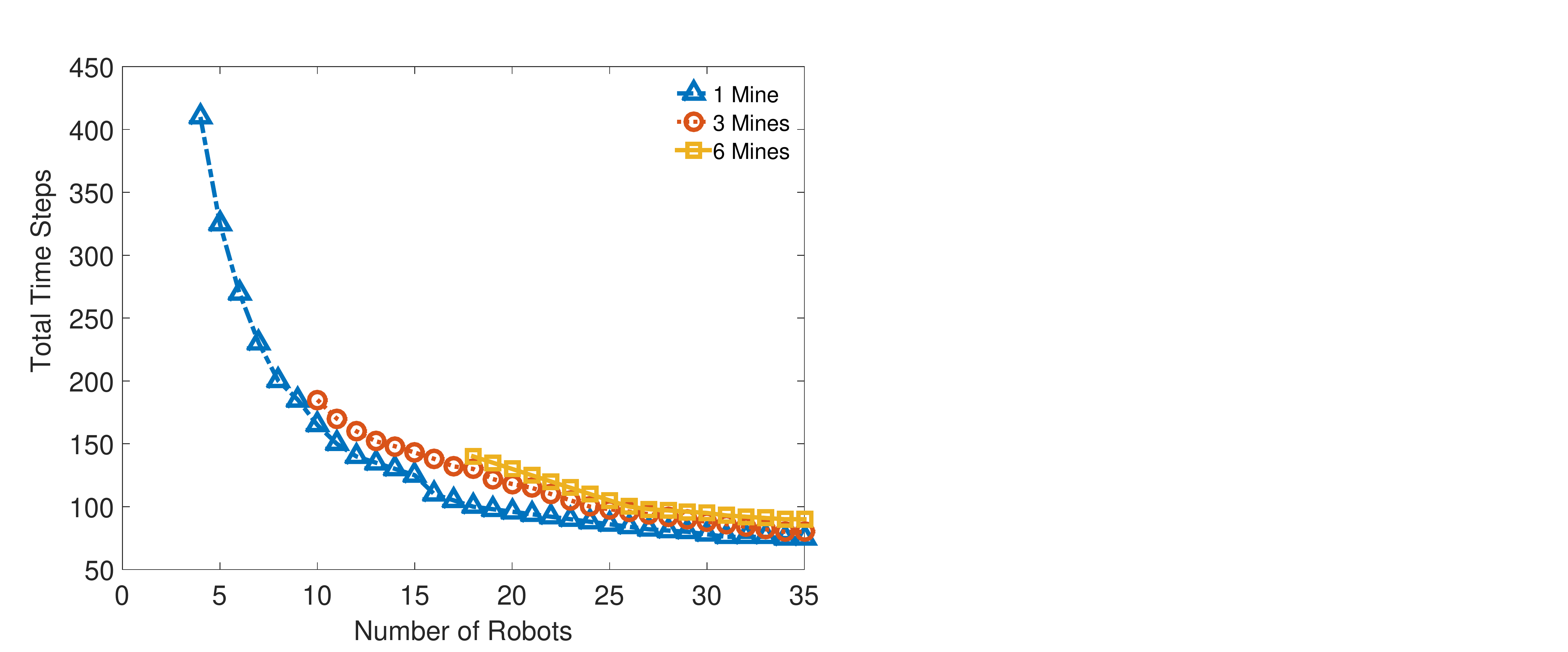}
\caption{Evaluation of the performance of ATRC in 30x30 grid area in terms of total time steps. }
	\label{fig:ProtocolVSMines}
\end{figure}

\begin{figure}
	\centering
	\includegraphics[scale=0.3]{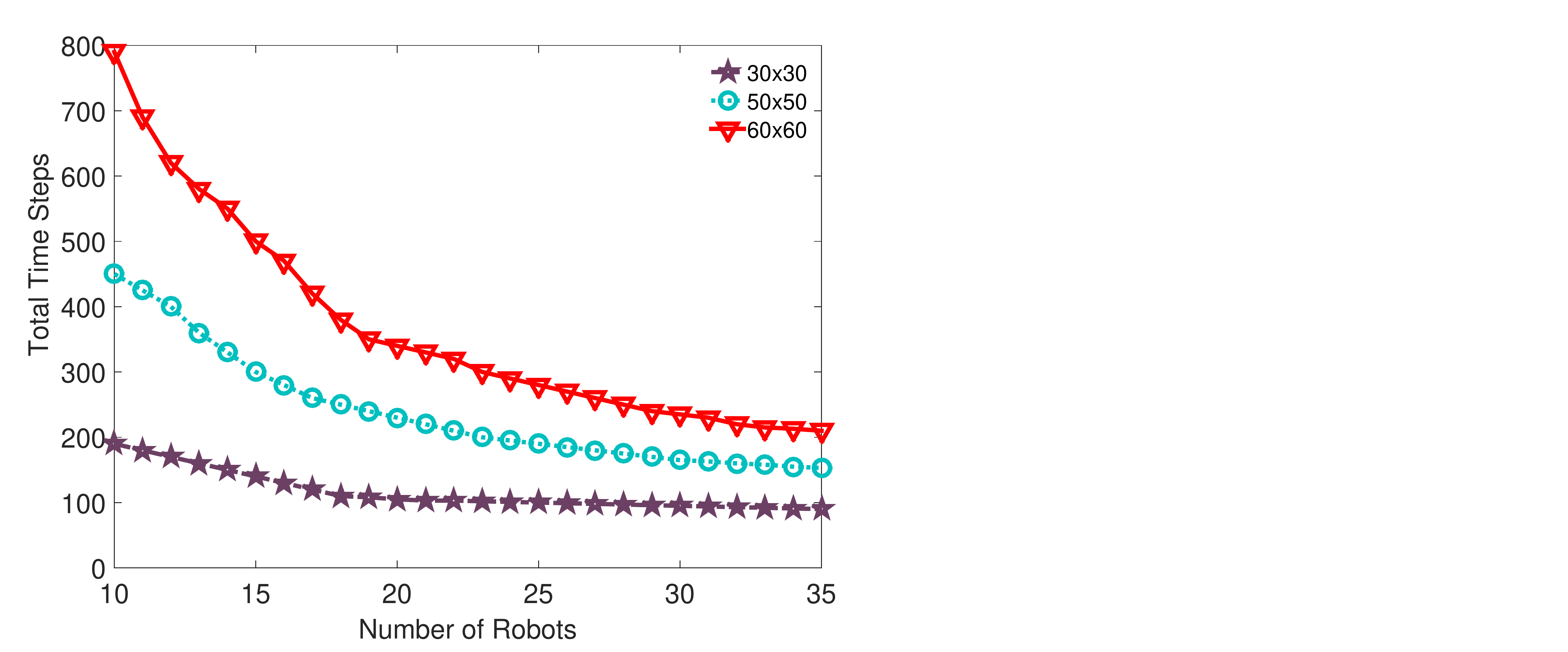}
	\caption{Evaluation of the influence of the dimension of the area in terms of  total Time Steps. }
	\label{fig:ProtocolVGrid}
\end{figure}

In Fig. \ref{fig:ProtocolVSMines} and Fig. \ref{fig:ProtocolVGrid} are shown the convergence time under
increasing number of mines and increasing grid size. It is
possible to see as the number of mines that can
increase the recruiting time and indirectly affect the
discovery time does not affect too much the overall
convergence time. This means that the ATRC is able to
dynamically adapt its strategy in recruiting and in the
discovery in order to maintain low the difference if the
complexity increases. Concerning Fig.~\ref{fig:ProtocolVGrid}, where the grid size increases, the ATRC-ERP increases the
convergence time for larger area. This is expected because
with the same number of robots it is necessary to take more time
to explore all the un-known area. In this case it is the
exploration time that affects the overall convergence time.
However, if the number of robots increases the
convergence time can be reduced and, after a certain
amount, having more robots do not introduce more any
benefits in the space discovery time.

\begin{figure}
	\centering
	\includegraphics[scale=0.3]{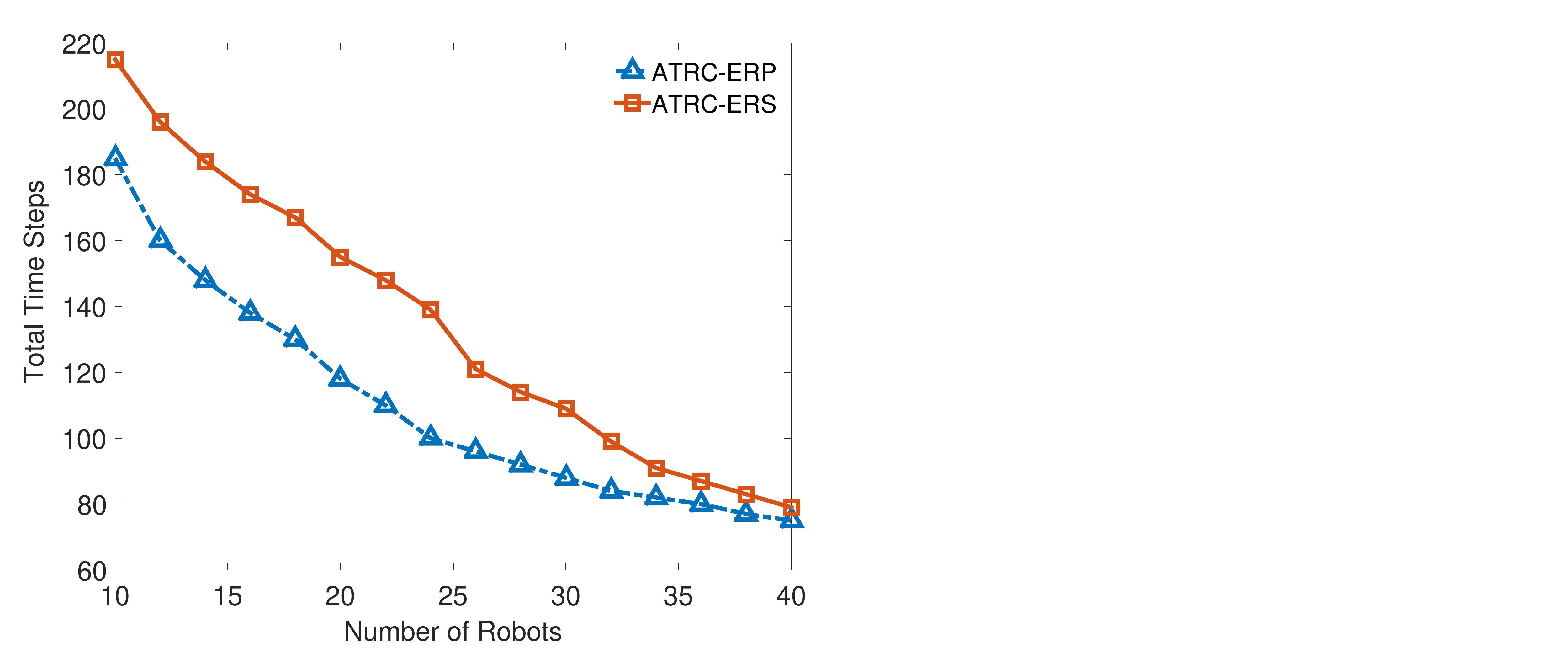}
	\caption{Comparison between ATRC-ERP and ATRC-ERS in a grid area 30x30 evaluating the total time steps. }
	\label{fig:StigmergyVSProtocol}
\end{figure}

In Fig. \ref{fig:StigmergyVSProtocol}, we compare the two proposed recruiting
strategies in a grid area 30x30 with 3 mines to disarm. It
can be seen as for a lower number of robots the wireless
communication (ATRC-ERP) performs better than the
mechanism with only stigmergy (ATRC-ERS) in terms of
number of iterations. This means that the communication
among the robots allows to complete the tasks (exploring
and recruiting/disarming) more quickly.
Increasing the size of swarm, the results are comparable
because the higher number of robots assures a natural
distribution among exploring and disarming tasks leading
to a reduced overall execution time.
Regarding the number of packets in Fig. 18 it is shown that
it mainly depends on the number of mines in the area. The
number of robots does not affect the overhead because the
proposed algorithm, such as designed, avoids an excessive
increase of packets forwarding in the network. The number
of packets in the network is nearly constant increasing the number of robots with a certain number of mines; instead
increasing the number of mines with a certain number of
robots the number of packets increases. This is due to the
scalable approach of ATRC that adopts just local
information to know where to send packets (highest link
selection probability) and global information through the
stigmergy avoiding to increase the control overhead to
maintain the robot topology and distribute tasks.

In Fig. \ref{fig:ProtocolGridVSPackets} it is shown the number of packets sent on the
network varying the grid area size. In this case the number
of packets increases proportionally to the size of area when
there are few robots because the network is instable and all
tracks cannot be completed and robots are not immediately
released to complete the exploration. However, the
network reaches the stability increasing the number of
robots and with the possibility to distribute both tasks
(recruiting and exploration) in the overall area.

	\begin{figure}
	\centering
	\includegraphics[scale=0.3]{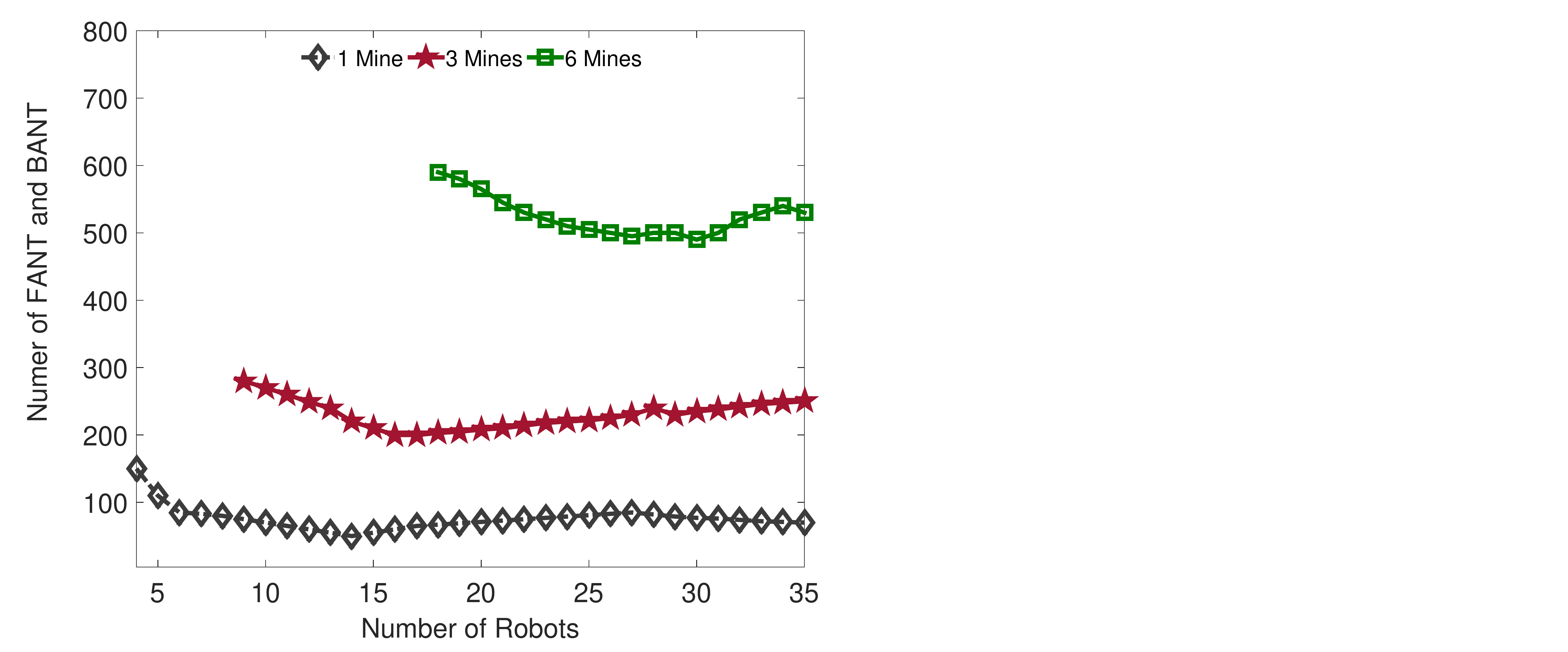}
	\caption{Evaluation of the influence of the number of mines on the performance of ATRC in 30x30 grid area in terms of number of sent packets. }
	\label{fig:ProtocolMINEVSPackets}
\end{figure}

	\begin{figure}
	\centering
	\includegraphics[scale=0.3]{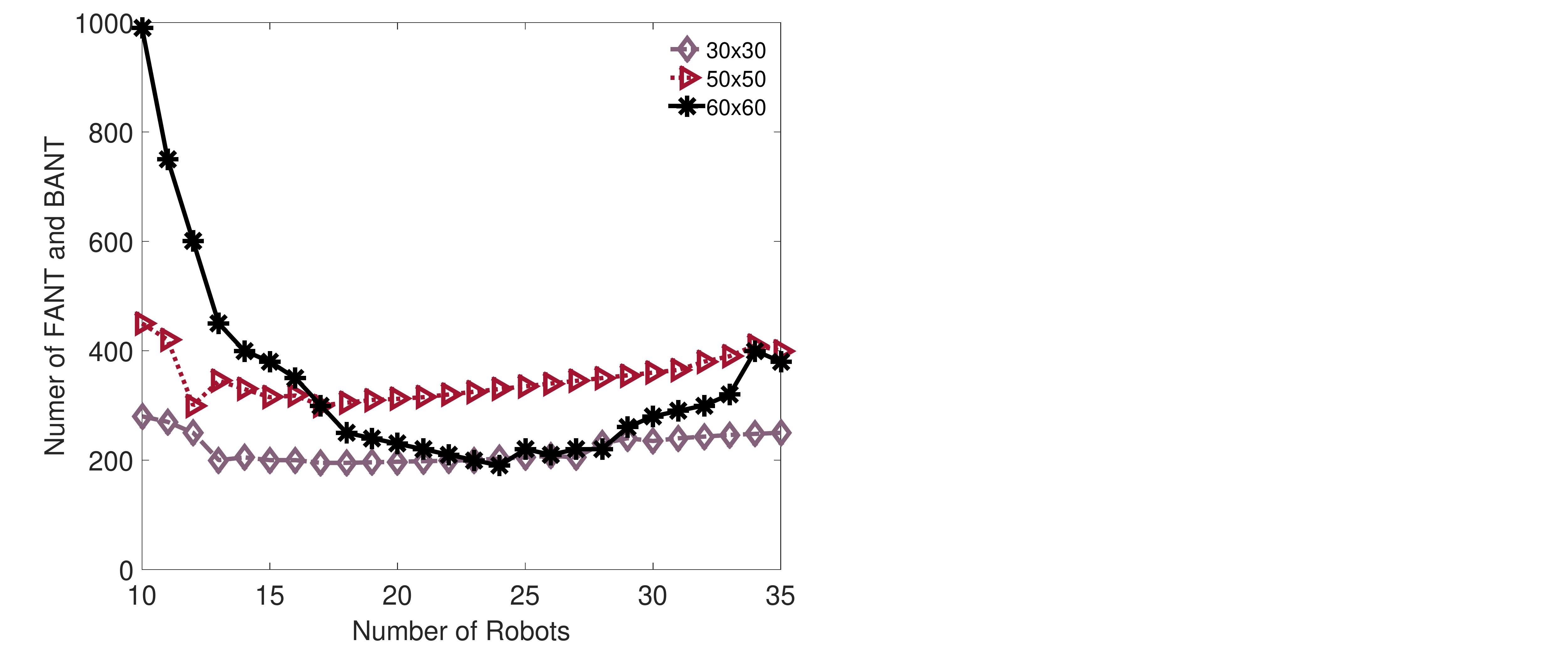}
	\caption{Evaluation of the influence of the dimension of the area on the performance of ATRC in terms of number of sent packets. }
	\label{fig:ProtocolGridVSPackets}
\end{figure}

\begin{table}
	\caption{Parameters used in the exploration algorithm. }
	\label{tab: explorationParameters}
	\begin{tabular}{cc}
		\hline\noalign{\smallskip}
		Parameters & Value  \\
		\noalign{\smallskip}\hline\noalign{\smallskip}
		Sensing range $R_s$  & 4 \\
		$\rho$ & 0.2  \\
		$\Delta\tau_0$ & 2 \\
		$\varphi$ & 1 \\
		$\lambda$ & 1 \\
		$\eta$ & 0.9\\
		$a_1$ & 0.5 \\
		$a_2$ & 0.5\\
		$\varepsilon$ & Uniform [0 1]\\
		\noalign{\smallskip}\hline
	\end{tabular}
\end{table}

\section{Conclusions}
In this paper, we have formulated a multiple task optimization problem for multiple mobile robots, and these main tasks are: the exploration of unknown area for detection mines and the recruitment for disarming them.
We have developed biologically inspired coordination strategies for robot swarms under complex constraints.
Based on the Ant Colony Algorithm, some modifications have been carried out to make these algorithms suitable for robot coordination and exploration tasks.

For the exploration task, we have used an indirect communication mechanism between the swarm based on the repelling anti
pheromone that tried to spread the robots in different regions of the area. For the recruitment task, we have proposed two strategies. The first is based on an indirect approach and uses an attractive pheromone to guide the swarm, the second uses a direct communication between the robots. For
this purpose, a new protocol able to disseminate recruiting requests and to recall the right number of robots to disarm mines in the minimal amount of time is presented. This protocol applies a probabilistic approach inherited by swarm-robotics in order to offer a scalable and distributed
solution to the mine disarming field issue. Such as verified by simulation results, our algorithm reduces the convergence time in comparison with IAS-SS. Moreover, the increase of the number of mines in the field lightly
increases the average convergence time while the increase of the research area (cells) lightly affects the system performance.
Self-organization of robots team with the addition of wireless communications to disseminate tasks and coordinate the robots (ATRC) reveal to be a good merging approach in the design of new kinds of protocols in this interesting research area.

Possible future works include the extension of methods to dynamically adjust the
number of hops to send the packets during the mission so as to be adaptive to the resource of the robots or other constraints. In addition, the proposed method can be modified to potentially deal with the unknown but mobile targets in an unknown area. Furthermore, further research can also consider the uncertainty concerning unreliable communication than can cause packets loss and inaccurate information, and thus make the overall system more reliable and robust.


\end{document}